\documentclass[letterpaper, 10 pt, conference]{ieeeconf}  

\IEEEoverridecommandlockouts                              

\usepackage{amsmath} 
\usepackage{graphicx}

\usepackage{balance}
\usepackage{booktabs}
\usepackage{multirow}
\usepackage{graphicx}

\usepackage{subcaption} 
\usepackage{stfloats}

\usepackage{algorithm}
\usepackage{amsmath} 
\usepackage{amssymb} 
\usepackage{flafter} 

\usepackage{xcolor}
\usepackage{xspace}
\usepackage{cite}

\usepackage{algpseudocode}
\usepackage{booktabs}    
\usepackage{multirow}   
\usepackage{graphicx}  
  
\usepackage{subcaption}  
\usepackage{stfloats}    
\usepackage[hidelinks,breaklinks]{hyperref}

\newcommand{\newtask}[2]{%
  \expandafter\DeclareRobustCommand\csname #1\endcsname{%
    \texttt{#2}}%
}
\newtask{Square}{Square}
\newtask{Can}{Can}
\newtask{BoxCleanup}{BoxCleanup}
\newtask{CanSort}{CanSort}
\newtask{Coffee}{Coffee}
\newtask{WoollyBall}{WoollyBallPnP}
\newtask{Handover}{PackageHandover}

\newcommand{\newmethod}[3]{%
  \expandafter\DeclareRobustCommand\csname #1\endcsname{%
    \textcolor[HTML]{#2}{#3}\xspace}%
}

\newmethod{methodname}{000000}{ResFiT} 

\let\OLDthebibliography\thebibliography
\renewcommand\thebibliography[1]{
  \OLDthebibliography{#1}
  \scriptsize
}

\usepackage[dvipsnames]{xcolor}
\newif\ifcomments
\commentstrue        

\title{\LARGE \bf
KoopmanFlow: Spectrally Decoupled Generative Control Policy via Koopman Structural Bias
}

\usepackage{hyperref}

\author{%
    Chengsi Yao\textsuperscript{*}$^{1,2,4}$, 
    Ge Wang\textsuperscript{*}$^{1,2,3}$, 
    Kai Kang\textsuperscript{*}$^{1,2,3}$, 
    Shenhao Yan$^{1,2}$, 
    Jiahao Yang$^{1,2}$, 
    Fan Feng$^{1,2}$, \\
    Honghao Cai$^{1,2,3}$, 
    Xianxian Zeng$^{4}$, 
    Rongjun Chen$^{4}$, 
    Yiming Zhao$^{1,2}$, 
    Yatong Han$^{1,2}$, 
    and Xi Li\textsuperscript{\dag}$^{1,2}$ \\[0.4em]
    \small $^{1}$ISing AI \\
    \small $^{2}$Future Network of Intelligence Institution, The Chinese University of Hong Kong, Shenzhen \\
    \small $^{3}$School of Science and Engineering, The Chinese University of Hong Kong, Shenzhen \\
    \small $^{4}$Guangdong Polytechnic Normal University 
}

\begin{document}

\begingroup
\renewcommand\thefootnote{\fnsymbol{footnote}}
\maketitle

\footnotetext[1]{These authors contributed equally to this work.}

\footnotetext[2]{Corresponding author.}
\endgroup

\thispagestyle{empty}
\pagestyle{empty}


\begin{abstract}
  Generative Control Policies (GCPs) show immense promise in robotic manipulation but struggle to simultaneously model stable global motions and high-frequency local corrections. While modern architectures effectively extract multi-scale spatial features, their underlying Probability Flow ODEs apply a uniform temporal integration schedule. This creates a critical representational bottleneck: compressed to a single step to satisfy real-time Receding Horizon Control (RHC) requirements, uniform ODE solvers mathematically smooth over sparse, high-frequency transients entangled within low-frequency steady states. To decouple these dynamics without accumulating pipelined errors, we introduce KoopmanFlow, a parameter-efficient generative policy guided by a Koopman-inspired structural inductive bias. Operating within a unified multimodal latent space infused with visual context, KoopmanFlow bifurcates generation at the terminal stage. Because visual conditioning occurs before spectral decomposition, both branches are inherently visually guided yet temporally specialized: a macroscopic branch structurally anchors slow-varying, visually-correlated trajectories via single-step Consistency Training, while a transient branch uses Flow Matching to isolate high-frequency residuals stimulated by sudden environmental visual cues (e.g., contacts or occlusions). Guided by an explicit spectral prior and optimized jointly via a novel asymmetric consistency objective, KoopmanFlow establishes a fused co-training mechanism. This allows the variant branch to naturally absorb localized visual dynamics without multi-stage error accumulation. Extensive experiments demonstrate KoopmanFlow significantly outperforms state-of-the-art baselines in contact-rich tasks requiring agile disturbance rejection. Trading a surplus latency buffer for a richer structural prior, KoopmanFlow achieves superior control fidelity and parameter efficiency while remaining within practical real-time deployment boundaries.

\end{abstract}
\begin{figure*}[t]\small\centering\includegraphics[width=\textwidth]{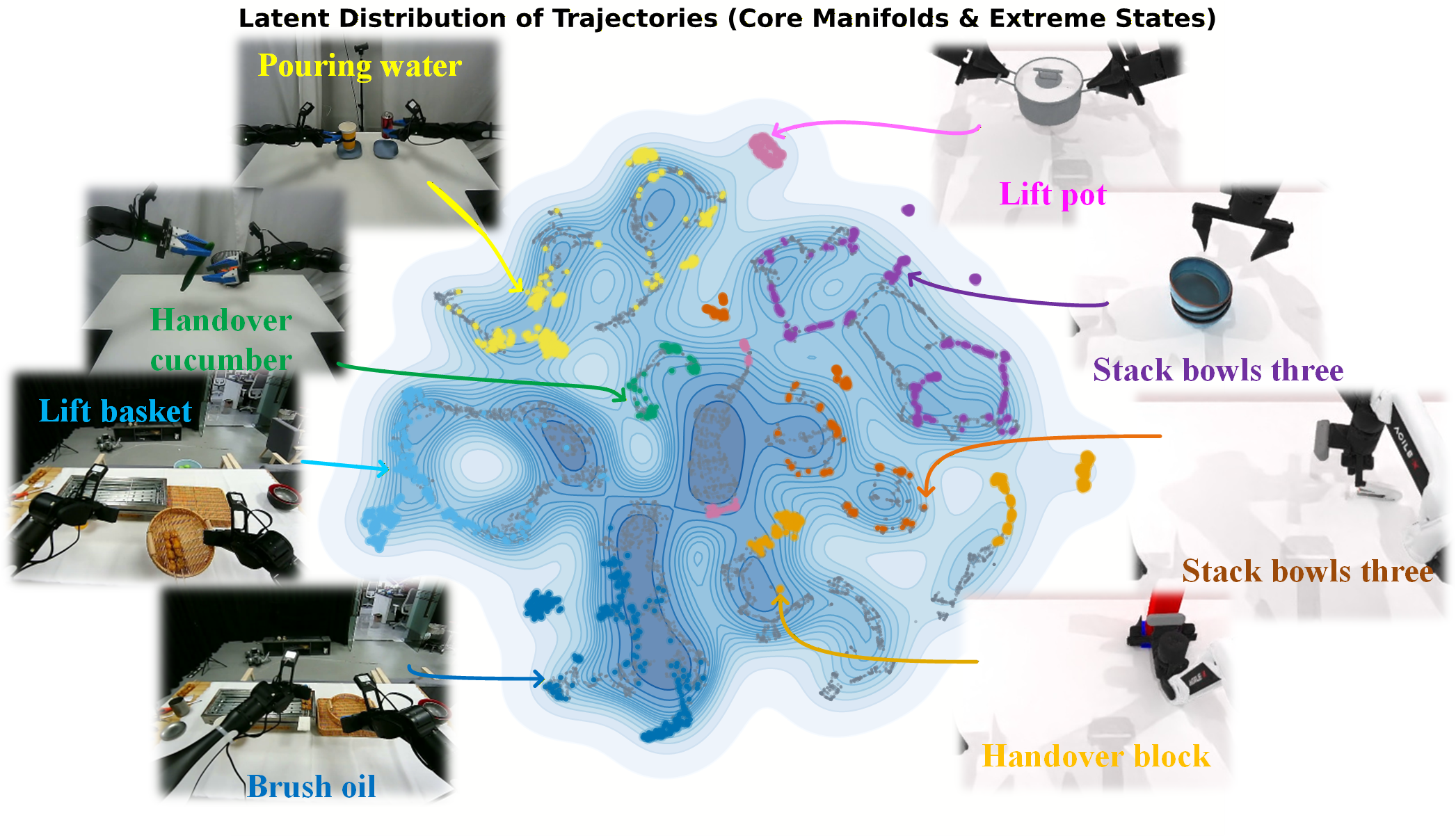}\caption{Latent Space Topography of Multi-Dataset Action Features. We visualize the action manifolds across multiple manipulation tasks, revealing severe topological overlap of distinct kinematic frequencies. Time-variant critical states (colored dots, representing grasping or contact) are deeply intertwined within the continuous trajectories of massive time-invariant steady states (grey dots). This severe spatial overlap highlights why standard continuous ODEs struggle: a single vector field cannot simultaneously resolve smooth inertia and high-frequency reactive corrections within the same topological neighborhood.}\label{fig:latent_space}\end{figure*}
\section{Introduction}\label{sec:intro}Recent breakthroughs in Generative Control Policies (GCPs), powered by Diffusion Models\cite{chi2023diffusion, zhu2024scaling, xu2025diffusion} and Flow Matching\cite{lipman2022flow, noh20253dflow}, have redefined the state-of-the-art in robotic visuomotor learning. However, deploying them in real-world, high-frequency control loops reveals a fundamental dilemma\cite{hirose2026asyncvla,tang2025vlash, black2025training}. Monolithic ODEs force a severe compromise: executing them for many steps (e.g., 10+ steps) captures high-fidelity reactive details but violates the strict latency bounds of Receding Horizon Control (RHC)\cite{duan2025real}; conversely, compressing them to a single inference step (NFE=1) via standard consistency distillation mathematically smooths over vital high-frequency corrections precisely when RHC needs them most\cite{prasad2024consistency, wang2024one}. This bottleneck stems not merely from the representation capacity of modern Transformer backbones, but from the uniform integration of the continuous vector field\cite{chun2025dynamic}. Standard ODE formulations force a single vector field to simultaneously resolve stable\cite{dong2025characterizing}, low-frequency global trends and stochastic, high-frequency local fluctuations\cite{zhong2025freqpolicy}.

This physical spectral heterogeneity manifests topographically as spectral confounding in the shared latent space\cite{zhong2025survey}. As illustrated in our latent space projection (Figure \ref{fig:latent_space}), time-variant critical states (e.g., exact grasping moments, denoted by colored dots) do not occupy isolated sub-manifolds\cite{julbe2025diffusion, jia2026action}. Rather, they are deeply intertwined within the continuous trajectories of massive time-invariant steady states (grey dots). This severe spatial overlap confounds monolithic ODE solvers, which struggle to abruptly shift from modeling stable inertia to predicting high-frequency corrections within the same topological neighborhood\cite{ting2025path}. Applying mathematical frequency filters to pure action sequences fails because the signals are kinematically fused. However, a critical breakthrough emerges when overlaying these representations with sensory stimuli: these high-frequency, entangled trajectories strictly co-occur with implicit semantic visual events (e.g., sudden occlusions or tool contacts)\cite{li2026causal,huang2026tic}. They are not stochastic noise, but visually-driven reactive corrections\cite{xie2026dynamicvla,wang2025vlatest}. To break this representational entanglement, we propose KoopmanFlow, a generative policy with a structural inductive bias mathematically inspired by Koopman operator theory ($\mathcal{K} \psi(x_{t}) = \psi(x_{t+1})$). Rather than forcing a single ODE trajectory to balance dense continuous cores and sparse topological fringes, KoopmanFlow structurally bifurcates the generative process at its terminal layer. Operating within a unified vision-proprioception latent space, visual context is injected prior to spectral decomposition. This allows a macroscopic branch, grounded by a physical Koopman reconstruction loss, to leverage Consistency Training to structurally anchor the low-frequency, visually-planned trajectory. Concurrently, a localized transient branch utilizes Flow Matching to isolate high-frequency residuals stimulated by sudden environmental visual cues. By collapsing the macro-level generation trajectory into a single step (NFE=1), KoopmanFlow offsets the micro-level computational overhead of exact Koopman operations. This design maintains practical real-time inference rates while introducing a richer structural prior, enabling high-fidelity control and robust contact handling without requiring massive parameter scaling or auxiliary spatial representations.

\section{Related Work}
\subsection{Generative Control Policies for Robotic Manipulation}
Continuous-time generative models, including Diffusion \cite{chi2023diffusion, zhu2024scaling, xu2025diffusion} and Flow Matching \cite{lipman2022flow, noh20253dflow}, excel in visuomotor imitation learning by effectively modeling highly multimodal action distributions \cite{ye2025radp}. To mitigate iterative sampling latency for real-world Receding Horizon Control (RHC), recent advances leverage Consistency Training \cite{song2023consistency, zhao2025hybrid} and architectural accelerations \cite{gong2025carp, yan2025maniflow} to enable few-step action generation without cumbersome distillation.

However, these accelerated frameworks typically apply a uniform integration schedule across the temporal action signal. By forcing a single continuous vector field to simultaneously resolve stable low-frequency trends and stochastic high-frequency fluctuations \cite{ye2025radp}, monolithic models suffer from spectral smoothing under 1-step RHC constraints, discarding vital reactive corrections. To address this, KoopmanFlow structurally decouples the action generation process based on the spectral heterogeneity of robotic dynamics. This paradigm avoids the pitfalls of optimizing solely for minimal latency, successfully balancing real-time RHC performance with the retention of high-fidelity, high-frequency control.

\subsection{Deep Koopman Dynamics and Spectral Decoupling}
Data-driven approximations of the Koopman operator, such as extended DMD (eDMD) \cite{koopman1931hamiltonian, schmid2010dynamic, lusch2018deep}, effectively linearize complex non-linear dynamics. While models like Koopa \cite{liu2023koopa} leverage this principle to disentangle time-invariant and time-variant components in discrete non-stationary time series, applying spectral decomposition to continuous-time generative control policies under diffusion noise remains unexplored.

KoopmanFlow bridges this gap via a fused consistency objective that geometrically constrains the macroscopic Koopman latent dynamics to a single-step mapping (NFE=1). This mechanism forces the unconstrained time-variant branch to naturally absorb visually-driven high-frequency residuals. Crucially, by collapsing the generative trajectory to a single macro-step, the architecture inherently offsets the computational overhead of exact DMD and prevents the error accumulation typical of multi-stage pipelines, ensuring robust viability for real-time deployment.

    \section{Methodology}
    To address the representational entanglement inherent in monolithic one-step generative policies, we introduce KoopmanFlow, a novel architecture that structurally decouples action generation through spectral filtering. As established in Section \ref{sec:intro}, high-frequency robotic corrections do not constitute random noise; rather, they are deterministic reactions driven by multi-modal stimuli (e.g., visual contacts, state shifts, or language cues). To bridge this multi-modal context with physical dynamics, the architecture establishes a visually conditioned latent space. This space is subsequently bifurcated into a time-invariant branch, which captures steady global inertia, and a time-variant branch, which absorbs visually stimulated reactive residuals.
    
    Instead of relying on a strict derivation from first principles, we introduce a structural prior inspired by the Koopman operator. This prior integrates Koopman operator theory, which models physical action dynamics, with Optimal Transport Flow Matching, which enables generative probability modeling.

      \begin{figure*}[t]
        \small
        \centering
        \includegraphics[width=0.8\textwidth]{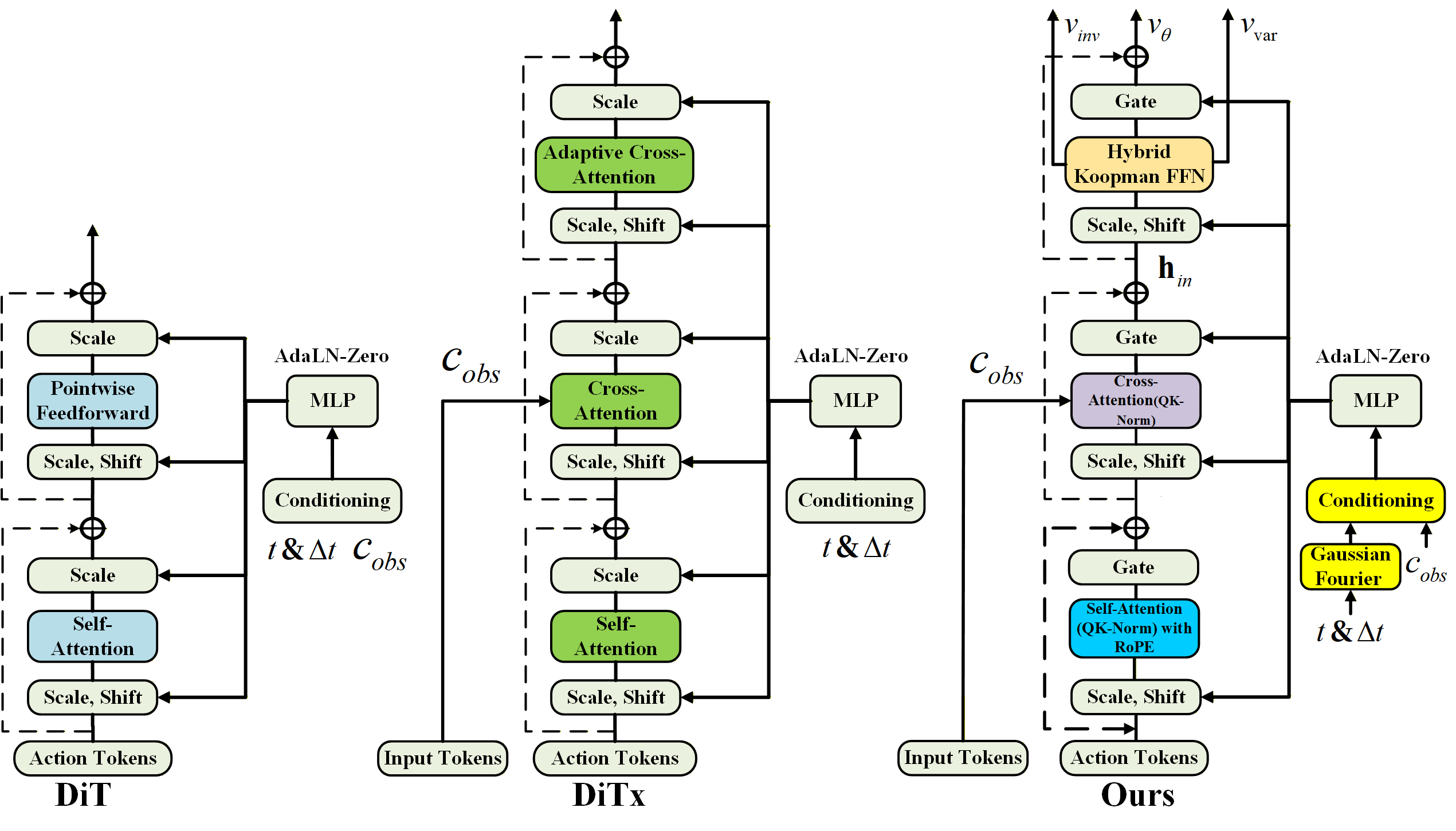}
        \caption{Architectural comparison of generative backbones. Left (DiT): The standard formulation using flattened observation conditioning. Middle (DiTx\cite{yan2025maniflow}): An architecture utilizing cascading cross-attention layers. Right (Ours): The proposed hierarchical condition injection, where the terminal Hybrid Koopman FFN decouples the velocity field into time-invariant ($\mathbf{v}_{inv}$) and time-variant ($\mathbf{v}_{var}$) dynamic components.}
        \label{fig:visual_events}
    \end{figure*}
    \subsection{Diffusion Transformer and Latent Spectral Decoupling}

    To map the theoretical dynamics into a generative model while maintaining high parameter efficiency, the generative backbone is built upon the Diffusion Transformer (DiT). As illustrated in Figure \ref{fig:visual_events}, we contrast with DiTx \cite{yan2025maniflow} by reintroducing observation conditioning ($c_{obs}$) into the AdaLN-Zero blocks, aligning closer to standard DiT. Specifically, $c_{obs}$ is explicitly constructed from visual and language information. The scalar diffusion timestep ($t$) and a Gaussian Fourier encoding of the event variable are fused with $c_{obs}$ to modulate AdaLN-Zero. Subsequently, the action tokens query dense multi-modal features via a QK-Norm-augmented cross-attention module, which is augmented with QK-Norm to bound the local Lipschitz constants.
    
    Crucially, these modality variables must fully transit through the DiT blocks to form the hidden representation $\mathbf{h}_{in} \in \mathbb{R}^{B \times T \times D}$ before entering the terminal Hybrid Koopman Feed-Forward Network (HKFFN). Disambiguating the temporal domains is crucial: the Koopman operator strictly models transitions along the physical sequence steps $\tau \in \{1, \dots, T\}$, whereas the diffusion time $t \in [0, 1]$ serves solely as a conditioning scalar defining the generative noise manifold.
    
    Because robotic contact dynamics are inherently non-stationary, global Fourier analysis is mathematically imprecise. To address this, an explicit structural prior is established along the physical sequence axis by applying a one-dimensional Real Fast Fourier Transform (RFFT). This transform serves as an efficient empirical proxy to split the latent space representing the entire trajectory:
    \begin{equation}
        \mathbf{X}_{inv}, \mathbf{X}_{var} = \text{FourierFilter}(\mathbf{h}_{in})
    \end{equation}
    By sorting the dataset-averaged amplitude spectra, the minimal set of top-amplitude frequencies that jointly accounts for a cumulative spectral energy threshold $\alpha$ is retained to form the time-invariant component sequence $\mathbf{X}_{inv}$.
    
    These spectrally decoupled latent trajectory features ($\mathbf{X}_{inv}, \mathbf{X}_{var}$) are processed through the respective physical operators, as mathematically detailed in Section 3.2:
    \begin{equation}
    \begin{aligned}
        \mathbf{h}_{inv} &= \mathcal{D}_{inv}(\mathcal{E}_{inv}(\mathbf{X}_{inv}) \tilde{\mathbf{K}}_{inv}) \\
        \mathbf{h}_{var} &= \mathcal{D}_{var}(\mathcal{E}_{var}(\mathbf{X}_{var}) \mathbf{K}_{loc}) 
    \end{aligned}
    \end{equation}
    
    Finally, these processed components bypass the standard residual connections to support the decoupled consistency routing. The components are independently projected to form sub-velocity fields before yielding the total predicted velocity $\mathbf{v}_\theta$:
    \begin{equation}
    \begin{aligned}
        \mathbf{v}_{inv} &= \text{Proj}_{inv}(\mathbf{h}_{inv}) \\
        \mathbf{v}_{var} &= \text{Proj}_{var}(\mathbf{h}_{var}) \\
        \mathbf{v}_\theta &= \mathbf{v}_{inv} + \mathbf{v}_{var}
    \end{aligned}
    \end{equation}

\subsection{Action Dynamics via Koopman Operator Theory}

After decoupling the visually-conditioned latent features, we formulate the mathematical operators that govern the corresponding physical transitions. In robotic manipulation, the temporal evolution of an action sequence constitutes a complex non-linear dynamical system across physical steps $\tau$. Let $\mathbf{a}_\tau \in \mathcal{M}$ denote the physical action at sequence step $\tau$. 

\textbf{Stochastic Koopman Operator for Time-Invariant Inertia:} Physically, the time-invariant branch models the robot's steady, predictable macroscopic trajectory. Inspired by the Wold decomposition theorem \cite{wold1938study}, we capture this smooth inertia using a global Koopman operator. By defining a measurement function $g$ and an encoder $\mathcal{E}_{inv}$, we map the physical state sequence into a high-dimensional space, approximating the global operator with a learnable matrix $\tilde{\mathbf{K}}_{inv}$ such that $g(\mathbf{a}_\tau) \tilde{\mathbf{K}}_{inv} \approx g(\mathbf{a}_{\tau+1})$.

However, Flow Matching applies additive Gaussian noise scaled by diffusion time $t$, producing a noisy full-sequence tensor $\mathbf{X}_t$. This superposition renders deterministic Koopman operators inadequate. To resolve this, we reframe the macroscopic transition by drawing upon the subspace decomposition from Koopa \cite{liu2023koopa} and using the Stochastic Koopman Operator (SKO) framework \cite{han2021desko} as a conceptual lens (detailed in Appendix A). Because architectural choices like QK-Norm lack strict global Lipschitz guarantees, we introduce a macroscopic consistency loss ($\mathcal{L}_{CT\_inv}$). Following the logic of isolating invariant dynamics \cite{liu2023koopa}, this loss penalizes velocity divergence across consecutive diffusion timesteps, acting as a structural regularizer to stabilize a stochastic-robust invariant subspace inspired by SKO principles. Consequently, $\tilde{\mathbf{K}}_{inv}$ robustly evolves the expected physical trajectory $\mathbb{E}[\mathbf{X}_t]$, successfully decoupling the macroscopic inertia from stochastic noise.

\textbf{Localized DMD for Time-Variant Residuals:} While the matrix $\tilde{\mathbf{K}}_{inv}$ governs the macroscopic inertia, robotic reactions involve non-stationary transient signals. To process these signals, the time-variant branch applies localized Dynamic Mode Decomposition ($\text{DMD}_{loc}$) over a dynamically scaled sliding window $\tau_w = \max(4, \lfloor T/4 \rfloor)$. 

By indexing the isolated time-variant sequence $\mathbf{X}_{var}$ along its physical steps, we extract the local state $\mathbf{x}_{var}^{(\tau)}$ to encode transient features as $\mathbf{Z}_\tau = \mathcal{E}_{var}(\mathbf{x}_{var}^{(\tau)}) \in \mathbb{R}^d$. By organizing these vectors as columns in consecutive state matrices $\mathbf{X} = [\mathbf{Z}_1, \dots, \mathbf{Z}_{\tau_w-1}]$ and $\mathbf{Y} = [\mathbf{Z}_2, \dots, \mathbf{Z}_{\tau_w}]$, where $\mathbf{X}, \mathbf{Y} \in \mathbb{R}^{d \times (\tau_w-1)}$, we compute the localized Koopman operator. To prevent degenerate manifolds ($\kappa(\mathbf{X}) \to \infty$) during physical singularities (e.g., collisions), we apply Tikhonov regularization with a damping factor $\lambda$:
\begin{equation}
    \mathbf{K}_{loc} = \mathbf{Y} \mathbf{X}^T (\mathbf{X} \mathbf{X}^T + \lambda \mathbf{I})^{-1}
    \label{eq:dmd_loc}
\end{equation}
This regularization bounds the norm of the operator without eliminating the transient signals, ensuring that visually-driven reactive corrections are preserved.

\textbf{Computational Viability:} Exact DMD in real-time control generally incurs a computational complexity of $\mathcal{O}(n^3)$. To address this issue, KoopmanFlow confines the exact DMD to the compressed latent bottleneck (\texttt{dynamic\_dim} $d=128$) structured by the HKFFN. Because the sliding window $\tau_w$ is small ($\tau_w \ll d$), the matrix $\mathbf{X}$ is tall and narrow. We compute the regularized pseudo-inverse via Truncated SVD ($\mathbf{X} = \mathbf{U}\mathbf{\Sigma}\mathbf{V}^T$), which resolves in microseconds. This approach enables exact transient spectral decomposition while remaining within the latency envelope of Receding Horizon Control ($<50$ ms).

\subsection{Flow Matching and Fused Co-Training}

\begin{figure*}[h]
    \centering
    \includegraphics[width=0.8\textwidth]{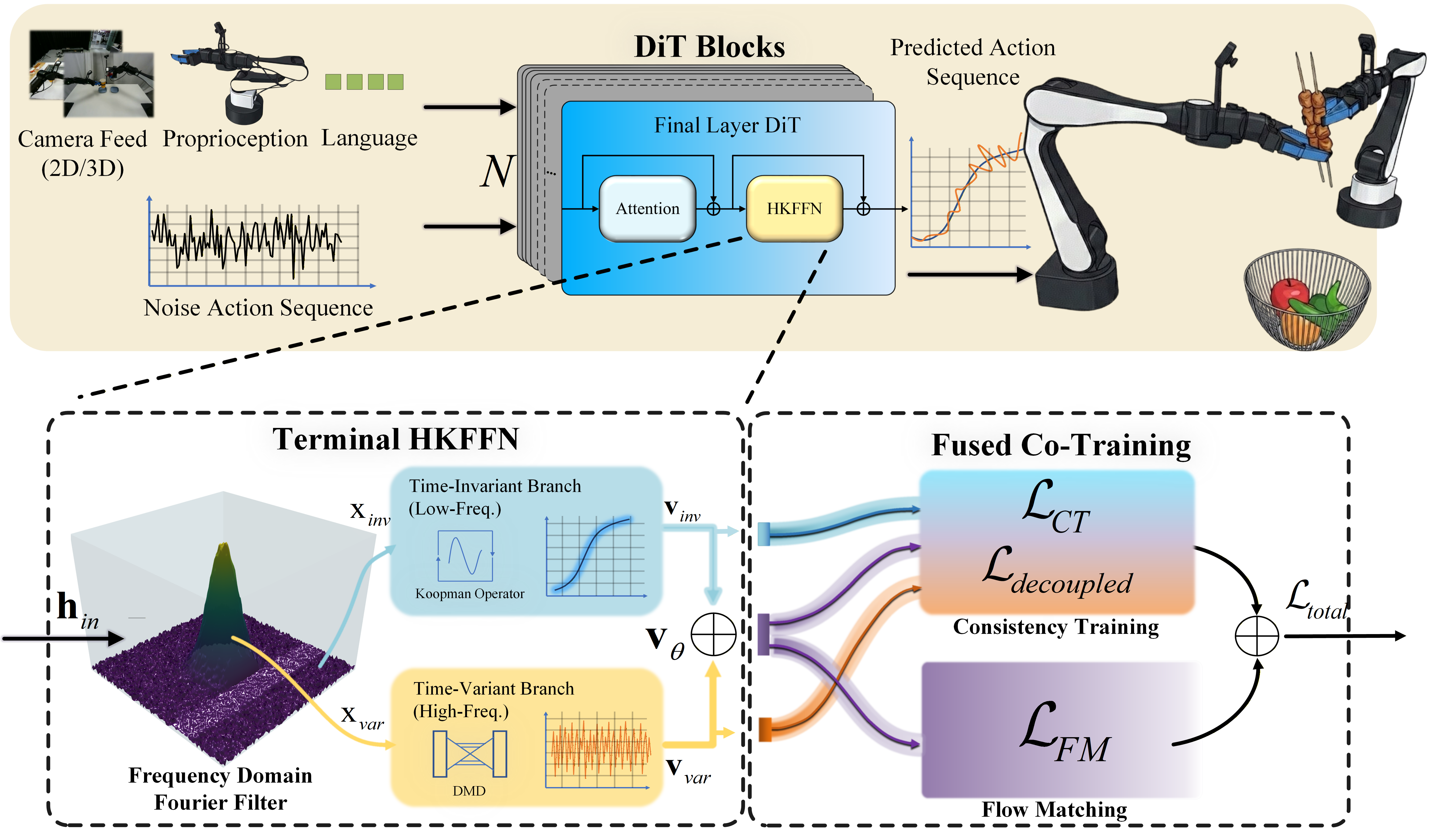} 
    \caption{KoopmanFlow architecture. Multi-modal features are processed via DiT and spectrally decoupled by a terminal HKFFN into time-invariant ($\mathbf{v}_{inv}$) and time-variant ($\mathbf{v}_{var}$) velocity fields. Fused Co-Training ensures exact spectral decomposition and single-step generation (NFE = 1) for real-time control.}
    \label{fig:KoopmanFlow_arch}
\end{figure*}

To reduce the number of function evaluations (NFE) during real-time deployment while preventing the representational smoothing typical of monolithic one-step models, KoopmanFlow combines Optimal Transport Flow Matching with the accelerated inference of consistency models \cite{song2023consistency, yan2025maniflow}. 

Enforcing Flow Matching alongside deep structural consistency distillation on the identical latent state frequently induces conflicting gradient signals. To ensure a stable gradient variance, we adopt the concurrent optimization strategy introduced by ManiFlow \cite{yan2025maniflow}. Instead of alternating objectives temporally, we asymmetrically partition each training batch spatially. We define a consistency batch ratio $r_{ct}$, where a fraction of the samples undergoes one-step consistency distillation and explicit spectral decoupling, while the remaining $(1 - r_{ct})$ fraction is dedicated to anchoring the continuous global trajectory via Conditional Flow Matching (CFM). In our main experiments, we set $r_{ct} = 0.2$. The total objective is formulated as follows:
\begin{equation}
    \mathcal{L}_{total} = \mathcal{L}_{FM} + \mathcal{L}_{CT} + \mathcal{L}_{decoupled} + \mathcal{L}_{reg}
\end{equation}

Operating over the generative diffusion time $t \in [0, 1]$, the flow-matching partition optimizes the standard CFM objective to construct a robust generative foundation for the entire physical sequence. To avoid ambiguity with the physical sequence steps $\tau$, let $\mathbf{X}_t \in \mathbb{R}^{T \times D}$ denote the full noisy trajectory over all physical steps at diffusion step $t$. The objective is:
\begin{equation}
    \mathcal{L}_{FM} = \mathbb{E}_{t, \mathbf{X}_0, \mathbf{X}_1} \left[ \left\| \mathbf{v}_\theta(\mathbf{X}_t, t) - (\mathbf{X}_1 - \mathbf{X}_0) \right\|^2_2 \right]
\end{equation}
where $\mathbf{X}_0 \sim \mathcal{N}(0, \mathbf{I})$ represents the pure noise sequence, $\mathbf{X}_1$ is the clean physical ground truth trajectory, and the vector field $\mathbf{v}_\theta$ models the generative flow derivative along $t$, entirely distinct from the physical robotic velocity along $\tau$.

For the remaining consistency partition, the model shifts to accelerated consistency distillation. Guided by a teacher network based on an exponential moving average (EMA), we construct a mathematically rigorous consistency flow target ($\mathbf{v}^{CT}_{target}$) for the entire trajectory:
\begin{equation}
\begin{aligned}
    \mathbf{v}^{CT}_{target} &= \frac{(\mathbf{X}_{t+\Delta t} + (1 - t - \Delta t) \mathbf{v}^{EMA}_{\theta}(\mathbf{X}_{t+\Delta t}, t+\Delta t)) - \mathbf{X}_t}{1-t} \\
    \mathcal{L}_{CT} &= \mathbb{E}_{t, \Delta t, \mathbf{X}_1} \left[ \left\| \mathbf{v}_\theta(\mathbf{X}_t, t) - \mathbf{v}^{CT}_{target} \right\|^2_2 \right]
\end{aligned}
\end{equation}

Crucially, within this consistency subset, the joint explicit and implicit decoupling ensures the one-step diffusion projection inherently respects the physical spectral boundaries formulated in Section 3.1:
\begin{equation}
    \mathcal{L}_{decoupled} = \lambda_{dec} \big( \mathcal{L}_{CT\_inv} + \mathcal{L}_{inv\_cross} + \mathcal{L}_{var\_flow} \big)
\end{equation}

\textbf{Theoretical Motivation for Unified Weighting:} Because the HKFFN utilizes a Fourier filter, the total signal energy is initially conserved across the frequency domain via the Parseval theorem. Consequently, we consider the sub-velocity fields ($\mathbf{v}_{inv}, \mathbf{v}_{var}$) to conceptually share a balanced latent energy space, and we adopt a unified weighting of $\lambda_{dec} = 0.5$ to optimally stabilize the gradient energy.

Within this decoupled objective, the term $\mathcal{L}_{CT\_inv}$ anchors the time-invariant trajectory at the current diffusion step, while $\mathcal{L}_{inv\_cross}$ rigorously enforces alignment across consecutive diffusion steps ($\Delta t$):
\begin{equation}
\begin{aligned}
    \mathcal{L}_{CT\_inv} &= \mathbb{E}_{t} \left[ \left\| \mathbf{v}_{inv}(\mathbf{X}_t, t) - \mathbf{v}_{inv}^{EMA}(\mathbf{X}_t, t) \right\|^2_2 \right] \\
    \mathcal{L}_{inv\_cross} &= \mathbb{E}_{t, \Delta t} \left[ \left\| \mathbf{v}_{inv}(\mathbf{X}_t, t) - \mathbf{v}_{inv}^{EMA}(\mathbf{X}_{t+\Delta t}, t+\Delta t) \right\|^2_2 \right]
\end{aligned}
\end{equation}
Simultaneously, the time-variant branch is implicitly routed to absorb the remaining physical transients. This process is explicitly guided by the semantic high-frequency predictions of the EMA teacher:
\begin{equation}
    \mathcal{L}_{var\_flow} = \mathbb{E}_{t} \left[ \left\| \mathbf{v}_{var}(\mathbf{X}_t, t) - \mathbf{v}_{var}^{EMA}(\mathbf{X}_t, t) \right\|^2_2 \right]
\end{equation}

By jointly optimizing the Optimal Transport flow with the consistency training objective, KoopmanFlow collapses the generation trajectory into a single step (NFE = 1). This reduction yields a favorable systemic trade-off: exchanging a large macro-level sequential bottleneck (iterative denoising) for a localized micro-level linear algebra operation. This synergy ensures that the architecture preserves exact spectral decomposition and high control fidelity for real-time deployment. To further enforce kinematic smoothness and prevent non-physical acceleration in the macroscopic trajectory, we apply an auxiliary spatiotemporal regularization term $\mathcal{L}_{reg}$ to the time-invariant branch during training. This constraint is strictly composed of $L_1$ penalties to effectively smooth persistent stochastic diffusion noise without compromising intentional macroscopic shifts.

\section{Experiment}

\begin{table*}[t] 
    \centering
    \caption{\textbf{Algorithm Performance Comparison on Adroit and DexArt Tasks.}}
    \label{tab:performance}
    \setlength{\tabcolsep}{5pt}
    \renewcommand{\arraystretch}{1.15} 
    \resizebox{\textwidth}{!}{ 
        \begin{tabular}{l l c c c c c c c c c}
            \toprule
            \multirow{2}{*}{\textbf{Algorithm}} & \multirow{2}{*}{\textbf{Obs.}} & \multicolumn{4}{c}{\textbf{Adroit}} & \multicolumn{5}{c}{\textbf{DexArt}} \\
            \cmidrule(lr){3-6} \cmidrule(lr){7-11}
             & & Hammer & Door & Pen & \textbf{Avg.} & Laptop & Faucet & Bucket & Toilet & \textbf{Avg.} \\
            \midrule
            \multicolumn{11}{l}{\textit{(Image-based)}} \\
            \addlinespace[2pt] 
            Diffusion Policy & Img & $54.0\pm3.6$ & $41.8\pm2.7$ & $18.5\pm2.5$ & $38.1\pm2.9$ & $81.7\pm2.1$ & $29.3\pm2.1$ & $26.0\pm2.4$ & $77.3\pm1.9$ & $53.6\pm2.1$ \\
            Flow Matching & Img & $55.7\pm4.2$ & $40.0\pm1.6$ & $21.2\pm0.8$ & $39.0\pm2.2$ & $81.7\pm2.5$ & $31.3\pm3.7$ & $24.0\pm2.2$ & $76.3\pm1.2$ & $53.3\pm2.4$ \\
            2D ManiFlow & Img & $100.0\pm0.0$ & $67.0\pm2.2$ & $56.0\pm3.6$ & $74.3\pm1.9$ & $85.7\pm2.1$ & $32.3\pm0.5$ & $29.7\pm3.4$ & $77.7\pm3.3$ & $56.3\pm2.3$ \\
            KoopmanFlow (Ours) & Img & $\mathbf{100.0\pm0.0}$ & $\mathbf{68.3\pm1.8}$ & $\mathbf{62.7\pm2.4}$ & $\mathbf{77.0\pm1.4}$ & $\mathbf{87.7\pm1.5}$ & $\mathbf{41.3\pm1.2}$ & $\mathbf{36.7\pm2.0}$ & $\mathbf{81.3\pm1.8}$ & $\mathbf{61.8\pm1.6}$ \\
            
            \midrule
            \multicolumn{11}{l}{\textit{(3D/PC-based)}} \\
            \addlinespace[2pt]
            3D Diffusion Policy & PC & $100.0\pm0.0$ & $76.7\pm4.7$ & $56.7\pm2.6$ & $77.8\pm2.4$ & $89.7\pm0.9$ & $41.7\pm0.5$ & $31.3\pm0.5$ & $79.7\pm0.9$ & $60.6\pm0.7$ \\
            3D Flow Matching & PC & $100.0\pm0.0$ & $77.7\pm6.1$ & $53.5\pm3.9$ & $77.1\pm3.3$ & $92.7\pm1.2$ & $42.0\pm0.8$ & $32.3\pm1.9$ & $79.7\pm0.5$ & $61.7\pm1.1$ \\
            3D ManiFlow & PC & $100.0\pm0.0$ & $80.3\pm1.2$ & $55.5\pm5.8$ & $78.6\pm2.3$ & $93.0\pm1.6$ & $45.0\pm3.6$ & $35.3\pm2.1$ & $79.3\pm3.3$ & $63.2\pm2.7$ \\
            KoopmanFlow (Ours) & PC & $\mathbf{100.0\pm0.0}$ & $\mathbf{86.3\pm1.5}$ & $\mathbf{69.3\pm2.2}$ & $\mathbf{85.2\pm1.2}$ & $\mathbf{98.3\pm0.8}$ & $\mathbf{60.7\pm1.5}$ & $\mathbf{67.7\pm1.8}$ & $\mathbf{89.7\pm1.2}$ & $\mathbf{79.1\pm1.3}$ \\
            \bottomrule
        \end{tabular}
    }
\end{table*}

\subsection{Experimental Setup}
\textbf{Benchmarks \& Baselines:} We evaluate KoopmanFlow across four diverse benchmarks: Adroit~\cite{rajeswaran2017learning} and DexArt~\cite{bao2023dexart} for high-DoF dexterous manipulation, MetaWorld~\cite{yu2020meta} (48 tasks) for language-conditioned multi-tasking, and RoboTwin2.0~\cite{chen2025robotwin} for complex bimanual coordination. We compare against state-of-the-art generative policies including Diffusion Policy (DP)~\cite{chi2023diffusion}, Flow Matching~\cite{lipman2022flow}, DP3~\cite{ze20243d}, $\pi_0$~\cite{physical2024pi0}, RDT~\cite{liu2024rdt}, and ManiFlow~\cite{yan2025maniflow} across 2D (RGB) and 3D (Point Cloud) modalities.

\begin{figure*}[t] 
    \begin{minipage}[b]{0.42\textwidth}
        \centering
        \captionof{table}{\textbf{1-Step Inference Performance (Laptop).}}
        \label{tab:1step_performance}
        \resizebox{\linewidth}{!}{
            \begin{tabular}{l c c c c}
                \toprule
                \textbf{Policy} & \textbf{Params} & \textbf{Latency} & \textbf{FPS} & \textbf{Acc} \\
                \midrule
                ManiFlow & $\sim$180M & 17.5 ms & 57.1 & $90.7 \pm 1.3\%$ \\
                DP3      & $\sim$255M     & 39.5 ms      & 25.3 & $86.7 \pm 2.1\%$ \\
                KoopmanFlow & $\sim$224M & 33.7 ms & 29.7  & $93.3 \pm 1.7\%$ \\
                \bottomrule
            \end{tabular}
        }
    \end{minipage}\hfill
    \begin{minipage}[b]{0.55\textwidth}
        \centering
        \captionof{table}{\textbf{Task Performance Comparison Across Different Models.}}
        \label{tab:model_comparison}
        \resizebox{\linewidth}{!}{
            \begin{tabular}{l c c c c c}
                \toprule
                \textbf{Task \textbackslash Model} & \textbf{DP} & \textbf{Pi0} & \textbf{DP3} & \textbf{RDT} & \textbf{KoopmanFlow} \\
                \midrule
                Adjust Bottle & $96\pm1.2\%$ & $90\pm5.0\%$  & $96\pm2.0\%$  & $81\pm3.0\%$  & $\mathbf{99\pm0.5\%}$ \\
                Press Stapler & $8\pm1.2\%$  & $62\pm5.0\%$  & $\mathbf{68\pm2.0\%}$  & $41\pm2.0\%$  & $67\pm3.2\%$ \\
                Lift Pot & $38\pm6.0\%$  & $84\pm3.0\%$  & $\mathbf{97\pm2.0\%}$  & $72\pm1.2\%$  & $81\pm4.2\%$  \\
                Stack Bowls 3 & $63\pm3.0\%$ & $66\pm1.1\%$ & $58\pm2.0\%$ & $56\pm3.0\%$ & $\mathbf{74\pm2.2\%}$ \\
                Handover Block & $45\pm7.0\%$  & $44\pm2.1\%$  & $70\pm3.0\%$  & $45\pm1.2\%$  & $\mathbf{81\pm2.1\%}$  \\
                \midrule
                \textbf{Average} & 50.0\%  & 69.2\%  & 77.8\%  & 59.0\%  & \textbf{80.4\%}  \\
                \bottomrule
            \end{tabular}
        }
    \end{minipage}

    \vspace{6mm} 

    \begin{subfigure}[b]{0.48\textwidth}
        \centering
        \includegraphics[width=\linewidth]{task_difficulty_comparison.png}
        \caption{\textbf{MetaWorld Results.}}
        \label{fig:realsuccess_rate_meta}
    \end{subfigure}
    \hfill 
    \begin{subfigure}[b]{0.48\textwidth}
        \centering
        \includegraphics[width=\linewidth]{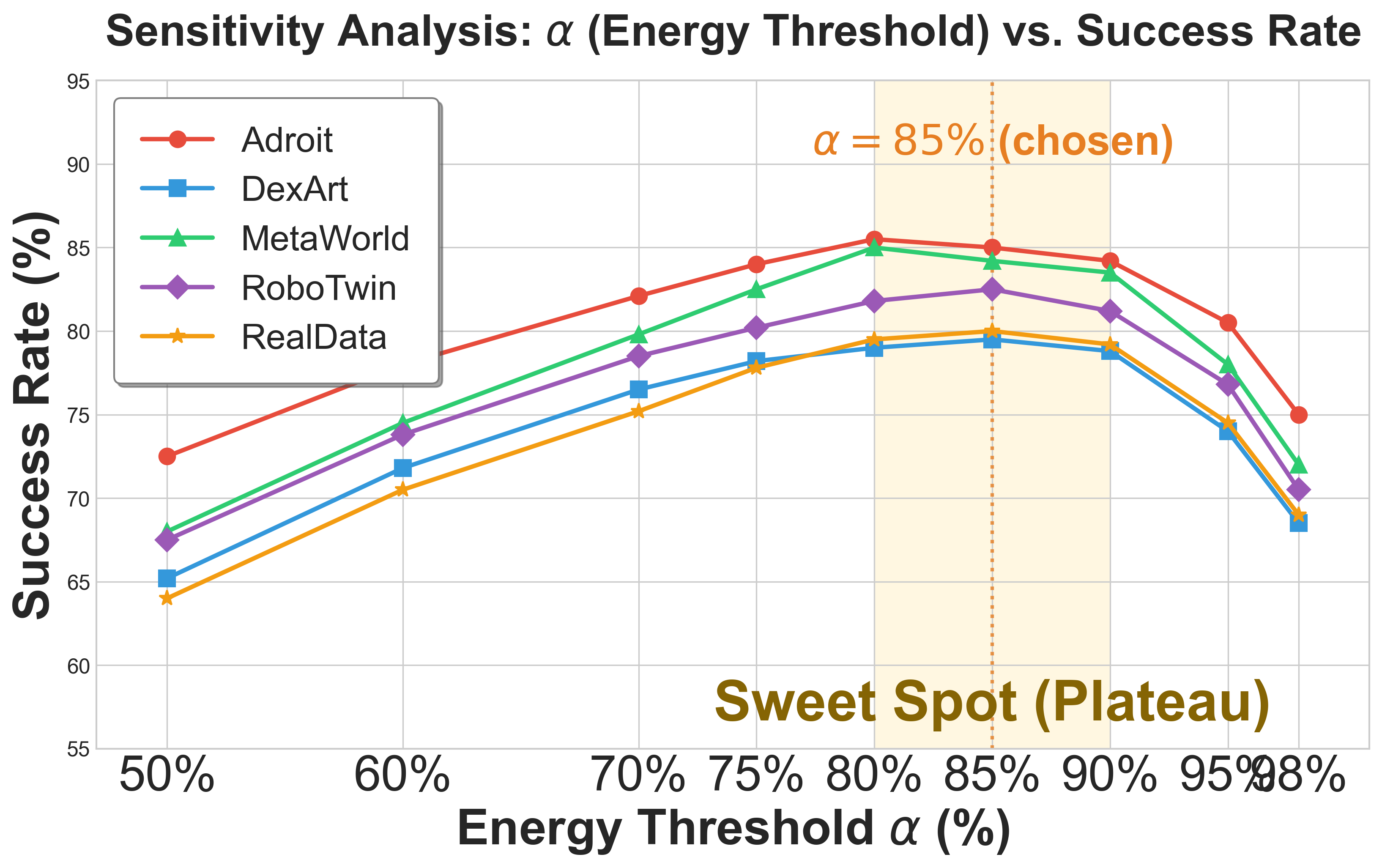}
        \caption{\textbf{Sensitivity Analysis of $\alpha$}.}
        \label{fig:sensitivity_alpha}
    \end{subfigure}
    
    \caption{\textbf{Experimental Analysis.} (Left) MetaWorld success rates across diverse tasks. (Right) Sensitivity analysis of the cumulative energy threshold $\alpha$. KoopmanFlow exhibits a robust performance plateau when $\alpha \in [80\%, 90\%]$, gracefully avoiding loss of global structure and high-frequency noise absorption.}
    \label{fig:combined_results}
\end{figure*}

\subsection{Main Results (Simulation)}
Tables~\ref{tab:performance}, \ref{tab:1step_performance}, \ref{tab:model_comparison}, and Figure~\ref{fig:combined_results} detail our comparative results. For language-conditioned tasks, we encode instructions into 512-dimensional semantic embeddings via a frozen T5 model, which are projected to guide the cross-attention mechanism. Overall, KoopmanFlow consistently outperforms continuous-time diffusion and flow matching baselines across both modalities.

\textbf{Dexterous \& Long-Horizon Manipulation}: In contact-rich tasks requiring continuous articulated interaction, 3D KoopmanFlow yields profound gains. Trajectory lengths reveal a stark dichotomy: stable tasks like \texttt{Laptop} and \texttt{Toilet} average 28.6 and 36.3 steps, where KoopmanFlow maintains near-perfect success (98.3\% and 89.7\%). However, tasks like \texttt{Faucet} and \texttt{Bucket} demand significantly longer horizons averaging 83.7 and 99.8 steps, making monolithic ODEs susceptible to compound error accumulation. Structurally anchoring the low-frequency trajectory via the macroscopic Koopman branch stabilizes sequence-level momentum, nearly doubling performance on the complex \texttt{Bucket} task (from 35.3\% in ManiFlow to 67.7\%).

\begin{table*}[t] 
    \renewcommand{\arraystretch}{1.2} 
    
    \begin{minipage}[t]{0.45\textwidth}
        \centering
        \caption{\textbf{Ablation of Penalty $\lambda_{dec}$.} Performance across Adroit under varying constraint rigidities.}
        \label{tab:lambda_ablation}
        \resizebox{\linewidth}{!}{
            \begin{tabular}{l c c c c c}
                \toprule
                \textbf{$\lambda_{dec}$} & \textbf{0.0} (None) & \textbf{0.1} & \textbf{0.5} & \textbf{1.0} & \textbf{2.0} (Over) \\
                \midrule
                Hammer & $82.3\pm4.2$ & $95.7\pm2.1$ & $\mathbf{100.0\pm0.0}$ & $98.3\pm1.2$ & $88.7\pm3.5$ \\ 
                \midrule 
                Door & $68.0\pm3.8$ & $77.3\pm3.2$ & $\mathbf{86.3\pm1.5}$ & $82.7\pm2.4$ & $76.3\pm3.1$ \\ 
                \midrule 
                Pen & $56.7\pm4.5$ & $61.0\pm3.1$ & $\mathbf{69.3\pm2.2}$ & $66.0\pm2.8$ & $59.0\pm4.0$ \\
                \midrule
                \textbf{Avg. (Adroit)} & $69.0\pm4.1$ & $78.0\pm2.8$ & $\mathbf{85.2\pm1.2}$ & $82.3\pm2.1$ & $74.7\pm3.5$ \\
                \bottomrule
            \end{tabular}
        }
    \end{minipage}\hfill
    \begin{minipage}[t]{0.53\textwidth}
        \centering
        \caption{\textbf{Consistency Batch Ratio ($r_{ct}$) Ablation.} Average Adroit success rates under varying $r_{ct}$.}
        \label{tab:routing_ablation}
        \resizebox{\linewidth}{!}{
            \begin{tabular}{l c c c c c c}
                \toprule
                $r_{ct}$ & 0.05 & 0.1 & \textbf{0.2} & 0.3 & 0.5 & 0.75 \\
                \midrule
                Hammer & $88.3\pm3.6$ & $96.0\pm1.8$ & $\mathbf{100.0\pm0.0}$ & $98.3\pm1.1$ & $85.7\pm4.0$ & $62.3\pm5.5$ \\ 
                \midrule 
                Door & $74.7\pm4.1$ & $82.3\pm2.6$ & $\mathbf{86.3\pm1.5}$ & $83.0\pm2.3$ & $71.3\pm3.8$ & $55.0\pm5.1$ \\ 
                \midrule 
                Pen & $60.0\pm4.3$ & $66.7\pm3.0$ & $\mathbf{69.3\pm2.2}$ & $65.7\pm2.7$ & $54.0\pm4.8$ & $41.7\pm5.9$ \\
                \midrule
                \textbf{Avg. (Adroit)} & $74.3\pm4.0$ & $81.7\pm2.4$ & $\mathbf{85.2\pm1.2}$ & $82.3\pm2.0$ & $70.3\pm4.2$ & $53.0\pm5.5$ \\
                \bottomrule
            \end{tabular}
        }
    \end{minipage}

    \vspace{5mm} 

    \begin{minipage}[t]{0.54\textwidth}
        \centering
        \caption{\textbf{Inference Step Ablation.} Efficiency vs. Performance on the DexArt benchmark.}
        \label{tab:inference_ablation}
        \resizebox{\linewidth}{!}{
            \begin{tabular}{l c c c c c c}
                \toprule
                \textbf{Algorithm} & \textbf{Infer Step} & \textbf{Laptop} & \textbf{Faucet} & \textbf{Bucket} & \textbf{Toilet} & \textbf{Avg.} \\
                \midrule
                3D ManiFlow & 10 & $93.0\pm1.6$ & $45.0\pm3.6$ & $35.3\pm2.1$ & $79.3\pm3.3$ & $63.2\pm2.7$ \\
                \midrule
                \multirow{4}{*}{3D KoopmanFlow} 
                 & 1  & $93.3\pm1.7$ & $46.7\pm3.4$ & $58.0\pm3.8$ & $\mathbf{89.7\pm1.2}$ & $71.9\pm2.6$ \\ \cmidrule(lr){2-7}
                 & 2  & $93.3\pm1.6$ & $48.0\pm3.1$ & $59.7\pm3.5$ & $83.3\pm2.6$ & $71.1\pm2.7$ \\ \cmidrule(lr){2-7}
                 & 5  & $95.7\pm1.2$ & $54.3\pm2.4$ & $\mathbf{67.7\pm1.8}$ & $81.0\pm2.8$ & $74.7\pm2.0$ \\ \cmidrule(lr){2-7}
                 & 10 & $\mathbf{98.3\pm0.8}$ & $\mathbf{60.7\pm1.5}$ & $65.3\pm2.2$ & $89.0\pm1.2$ & $\mathbf{78.3\pm1.4}$ \\
                \bottomrule
            \end{tabular}
        }
    \end{minipage}\hfill
    \begin{minipage}[t]{0.44\textwidth}
        \centering
        \caption{\textbf{Loss Ablation.} Optimization objectives vs. baseline 3D ManiFlow.}
        \label{tab:architecture_ablation}
        \resizebox{\linewidth}{!}{
            \begin{tabular}{l l c c c}
                \toprule
                \textbf{Algorithm} & \textbf{Variant} & \textbf{Door} & \textbf{Faucet} & \textbf{Bucket} \\
                \midrule
                3D ManiFlow & - & $80.3\pm1.2$ & $45.0\pm3.6$ & $35.3\pm2.1$ \\
                \midrule
                \multirow{4}{*}{3D KoopmanFlow} 
                 & w/o $\mathcal{L}_{CT}$         & $78.3\pm4.2$ & $43.0\pm3.8$ & $55.7\pm4.5$ \\ \cmidrule(lr){2-5} 
                 & w/o $\mathcal{L}_{decoupled}$  & $80.7\pm3.1$ & $52.3\pm3.2$ & $60.0\pm3.6$ \\ \cmidrule(lr){2-5}
                 & w/o $\mathcal{L}_{reg}$        & $84.3\pm2.3$ & $56.0\pm2.4$ & $65.7\pm2.7$ \\ \cmidrule(lr){2-5}
                 & Full Model                     & $\mathbf{86.3\pm1.5}$ & $\mathbf{60.7\pm1.5}$ & $\mathbf{67.7\pm1.8}$ \\
                \bottomrule
            \end{tabular}
        }
    \end{minipage}
    
    \vspace{6mm} 

    \begin{minipage}[t]{\textwidth}
        \centering
        \includegraphics[width=0.8\textwidth]{arch.png} 
        \captionof{figure}{\textbf{Internal Ablation Studies} on the Handover Block task.}
        \label{fig:internal_ablations}
    \end{minipage}
    
\end{table*}
\textbf{Multi-Task Generalization:} On MetaWorld (Figure~\ref{fig:combined_results}(a)), KoopmanFlow achieves $84.7\%$ overall success. In Hard/Very Hard tiers, monolithic models often suffer representational collapse from visual-semantic entanglement. Mitigating this spectral confounding, KoopmanFlow significantly outperforms 3D ManiFlow on unsaturated tasks (e.g., \texttt{Dial Turn}: $96\%$ vs. $67\%$) while maintaining saturated performance ($100\%$) on simpler ones.

\textbf{Bimanual Coordination \& Efficiency:} Evaluated on RoboTwin2.0 (Table~\ref{tab:model_comparison}), integrating the Hybrid Koopman FFN across all 8 DiT blocks allows our 2D RGB-only KoopmanFlow to achieve $80.4\%$ success, surpassing depth-conditioned 3D baselines like DP3 ($77.8\%$). Notably, in the contact-heavy \texttt{Handover Block}, our RGB model reaches $81\%$ (vs. DP3's $70\%$). Crucially, as detailed in Table~\ref{tab:1step_performance}, while the Koopman branch introduces a slight complexity overhead compared to base ManiFlow, KoopmanFlow outpaces DP3 in inference efficiency. It sustains a real-time 30 FPS execution rate while securing state-of-the-art success rates, confirming that structural advantages extend beyond 3D representations without compromising deployment viability.

\subsection{Inference Efficiency \& Ablation Studies}
\textbf{High-Fidelity 1-Step Control}: KoopmanFlow sustains exceptional performance under extreme low-latency constraints. Table~\ref{tab:inference_ablation} shows a 1-step KoopmanFlow achieves $89.7\%$ success on \texttt{Toilet}, decisively outperforming fully expanded 10-step ManiFlow ($79.3\%$). This confirms macroscopic consistency distillation captures core physical dynamics without catastrophic representational smoothing during aggressive step reduction.

\begin{figure*}[t]
    \centering
    \includegraphics[width=\textwidth]{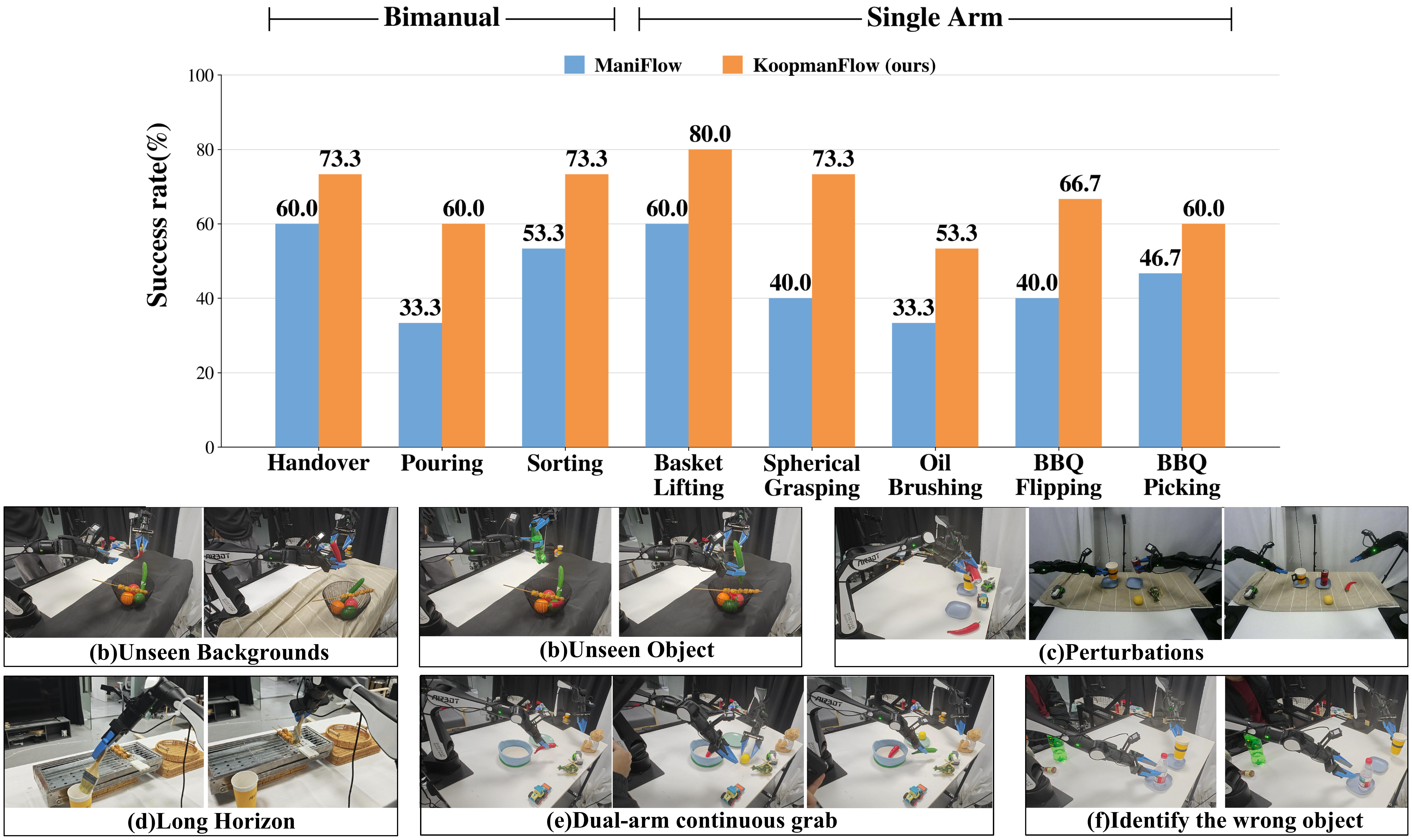} 
    \caption{\textbf{Quantitative success rate comparison between ManiFlow and KoopmanFlow} across eight real-world manipulation tasks. To maintain fairness, both models utilize a frozen R3M (ResNet-18) visual encoder from a single-camera view and are trained on 65 episodes per task. Results are averaged over 30 real-world trials under varied environmental conditions using 1 and 10 inference steps. KoopmanFlow exhibits superior performance, notably in long-horizon tasks requiring continuous TPU gripper coordination.}
    \label{fig:realsuccess_rate_main}
\end{figure*}

\begin{figure*}[t] 
    \centering
    \includegraphics[width=\textwidth]{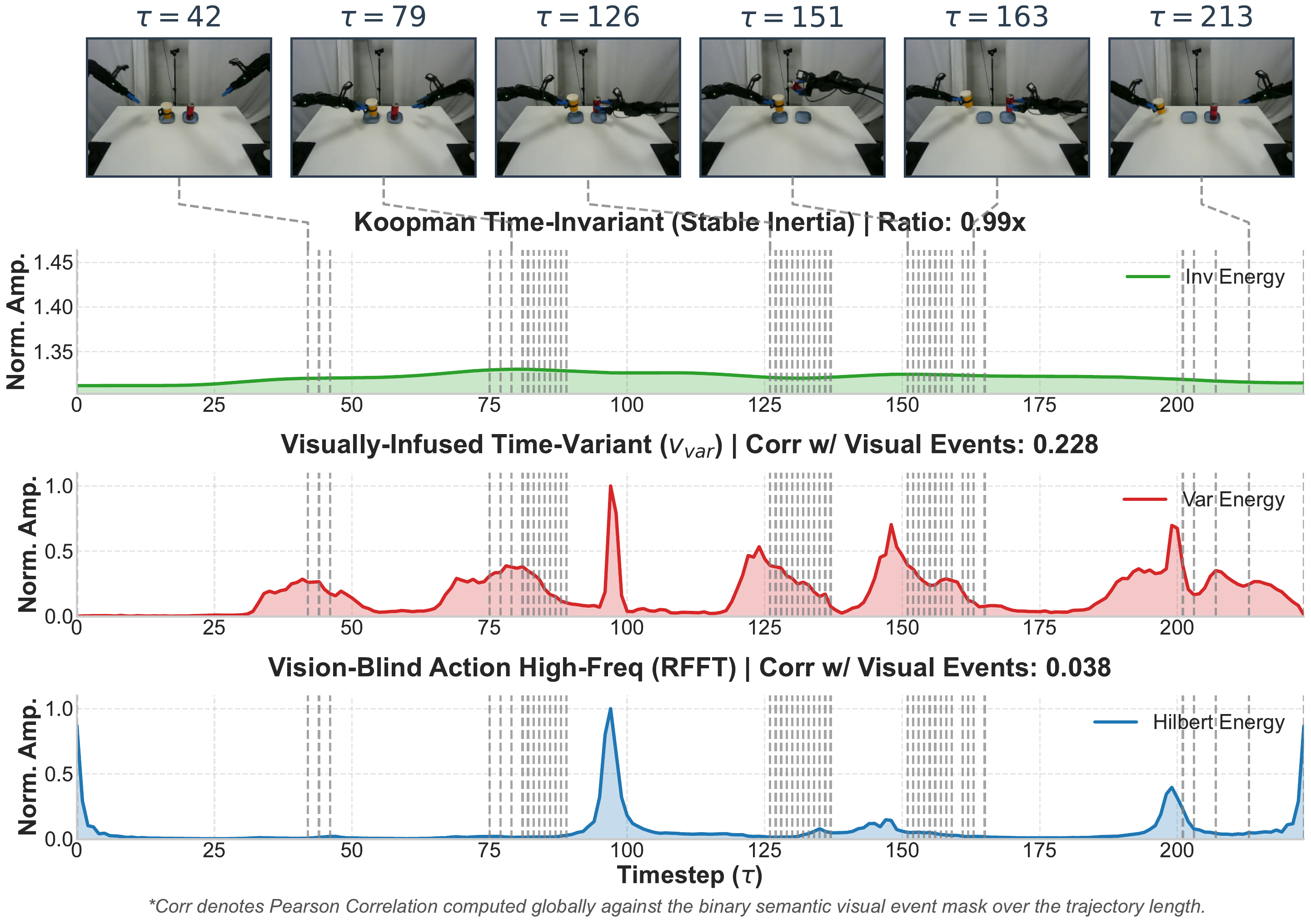} 
    \caption{\textbf{Qualitative visualization of the decomposed latent dynamics in the Hybrid Koopman FFN.} \textbf{(Top)} The time-invariant operator ($\mathbf{v}_{inv}$) maintains consistently stable absolute energy across the trajectory, capturing physical inertia unperturbed by sudden visual events. \textbf{(Middle)} The visually-conditioned time-variant operator ($\mathbf{v}_{var}$) exhibits sharp energy spikes that are highly correlated with semantic visual events (dashed lines), acting as a vision-driven reactive module to environmental disruptions. \textbf{(Bottom)} Naive high-frequency energy extracted via RFFT on raw action data lacks visual context and fails to align with these critical semantic events. This contrast demonstrates that KoopmanFlow achieves true semantic spectral decoupling rather than blind mathematical filtering.}
    \label{fig:koopffn}
\end{figure*}

\textbf{Architecture \& Objective Ablations}: For the decoupling penalty $\lambda_{dec}$ (Table~\ref{tab:lambda_ablation}), theoretical structural constraints align with empirical results. When $\lambda_{dec} = 0.0$, the structural constraint collapses causing spectral entanglement. Conversely, an excessive $\lambda_{dec} = 2.0$ causes the model to over-fit isolated sub-manifolds at the expense of the global closed vector field. Optimal stability emerges at $\lambda_{dec} = 0.5$. Regarding consistency batch ratio $r_{ct}$ and energy $\alpha$ thresholds, KoopmanFlow's gains stem from its decoupled architecture, not brittle tuning. The model shows a stable performance plateau for $r_{ct} \in [0.1, 0.3]$, validating our asymmetric spatial batch partition strategy. For $\alpha$ (Figure~\ref{fig:combined_results}(b)), KoopmanFlow maintains near-optimal performance between $80\%$ and $90\%$. If $\alpha < 75\%$, the variant branch models inertia, elegantly acting as a dynamic fallback to prevent total policy collapse at sub-optimal thresholds. Analyzing loss components (Table~\ref{tab:architecture_ablation}), the consistency loss is vital. Without $\mathcal{L}_{CT}$, the time-invariant branch ($v_{inv}$) absorbs stochastic diffusion noise rather than maintaining stability, causing severe unphysical jittering that drops \texttt{Faucet} performance from $60.7\%$ to $43\%$. Removing explicit decoupling (w/o $\mathcal{L}_{decoupled}$) or spatiotemporal kinematic regularization (w/o $\mathcal{L}_{reg}$) similarly degrades performance, confirming precise residual labor division is essential. Finally, for the generative backbone and HKFFN, the Hybrid Koopman FFN acts as a critical structural prior. Figure~\ref{fig:internal_ablations} shows that on the Handover task, reverting to a standard entangled MLP readout collapses success from $81.0\%$ to $31.5\%$. Crucially, HKFFN prevents catastrophic overfitting in extreme low-data regimes (10 to 20 demos) by successfully isolating stable physical dynamics. Furthermore, backbone stabilizers are vital: ablating QK-Norm/RoPE degrades performance to $76.8\%$, and replacing Gaussian Fourier conditioning drops it to $78.2\%$, confirming their respective roles in bounding attention variance and smoothing the diffusion manifold.


\subsection{Real-World Experiments \& Latent Visualization}

\textbf{Hardware Setup \& Evaluation Protocol}: To evaluate our approach physically, we deploy KoopmanFlow across two distinct hardware setups: the AIRBOT TOK4 mobile manipulation platform and the PTK4 fixed platform, evaluating a mix of single-arm operations and bimanual continuous tasks. The evaluation encompasses a diverse suite of eight physical manipulation tasks (Fig.~\ref{fig:realsuccess_rate_main}), designed to stress-test dexterous, long-horizon sequence modeling using TPU grippers. To systematically assess policy robustness and generalization, our experimental protocol introduces multiple axes of environmental variance, including altered backgrounds, distinct target object geometries, the deliberate introduction of visual distractors, varying grasp orientations, and extended bimanual coordination phases.

\textbf{Comparative Performance}: For a fair comparison between KoopmanFlow and the ManiFlow baseline, all evaluated policies operate from a single-camera viewpoint and process visual observations through a frozen R3M (ResNet-18) encoder. Furthermore, the models are trained in a data-scarce regime using 65 expert demonstrations per task. To test execution fidelity under real-time receding horizon limits, we constrain inference to extreme computational budgets of 1 and 10 Number of Function Evaluations (NFEs). Averaged over 30 real-world trials per task across 3 random seeds, KoopmanFlow consistently outperforms the baseline. The performance margin is most pronounced in prolonged, contact-rich TPU gripper manipulations, where KoopmanFlow successfully resists the compound error accumulation causing baseline monolithic models to fail.

\textbf{Latent Dynamics Analysis}: To explain this physical resilience, we visualize the decomposed latent dynamics of our Hybrid Koopman FFN (Fig.~\ref{fig:koopffn}). The time-invariant operator ($\mathbf{v}_{inv}$) maintains stable feature energy across the trajectory (Event/Away ratio: $0.99\times$), cleanly isolating physical inertia and macroscopic momentum unperturbed by sudden visual disruptions. Conversely, the time-variant operator ($\mathbf{v}_{var}$) exhibits sharp frame-to-frame spikes strictly aligned with semantic visual events (e.g., abrupt state changes during grasping). Crucially, while a naive mathematical high-pass filter (RFFT) on pure kinematic data fails to align with these semantic visual cues (Pearson correlation of $0.038$), our visually-infused $\mathbf{v}_{var}$ increases this correlation to $0.228$---a six-fold improvement. This contrast proves KoopmanFlow does not merely apply a blind mathematical frequency filter to actions, but implements a true vision-driven routing mechanism enabling precise, event-triggered reactive corrections during complex real-world execution.

\section{Conclusion and Limitations}
We presented KoopmanFlow to mitigate spectral entanglement in generative robotic control. By explicitly decoupling spectra during action generation, our approach significantly enhances real-time control fidelity. However, limitations remain. Relying on visual cues for reactive corrections makes the system susceptible to severe occlusions, which can suppress transient responses ($v_{var}$). Furthermore, the 1D RFFT may struggle to isolate transients during extreme, unmodeled physical shocks. Overall, KoopmanFlow demonstrates the potential of spectral decoupling as a robust paradigm for advanced action modeling.

\balance






\bibliographystyle{IEEEtran}
\bibliography{references}  

\clearpage
\onecolumn
\raggedbottom
\renewcommand{\baselinestretch}{1.1}
\appendix
\section*{Appendix A: Theoretical Motivation and Empirical Justification of Koopman-Inspired Dynamics on Flow Matching States}
\label{app:koopman_justification}

A valid concern regarding our Hybrid Koopman FFN is the application of the deterministic Koopman operator on intermediate states $\mathbf{x}_t$ corrupted by Gaussian noise $\mathbf{x}_0$. Rather than asserting a flawless mathematical equivalence under highly stochastic conditions, we strictly frame our approach as a \textbf{Koopman-inspired structural prior}. Here, we demonstrate how the specific geometry of Optimal Transport Flow Matching (OT-FM), combined with our architectural conditioning and empirical consistency regularization, conceptually motivates and practically bounds the error on these linear dynamics.

\textbf{1. The Deterministic Nature of PF-ODE}

Unlike traditional diffusion models governed by Stochastic Differential Equations (SDEs), OT-FM defines a deterministic Probability Flow ODE:
$$d\mathbf{x}_t = \mathbf{v}_\theta(\mathbf{x}_t, t) dt$$
Since the generation path is a deterministic flow mapping the standard Gaussian base to the target physical distribution, the system behaves as a non-autonomous deterministic dynamical system conditioned on $t$.

\textbf{2. Koopman Operator on Linear Interpolants and Local Lipschitz Conditioning}

Under the OT formulation, the intermediate state is strictly a linear combination: $\mathbf{x}_t = t \mathbf{x}_1 + (1-t) \mathbf{x}_0$. Let $\mathcal{E}$ be our observation (encoder) function lifting the state into the Koopman invariant subspace, and $\mathbf{K}$ be the learned Koopman matrix modeling the physical transition $\tau \rightarrow \tau+1$. For the ideal clean state $\mathbf{x}_1$, the Koopman evolution perfectly satisfies:
$$\mathcal{E}(\mathbf{x}_1^{\tau+1}) = \mathbf{K} \mathcal{E}(\mathbf{x}_1^\tau)$$
When applied to the noisy intermediate state $\mathbf{x}_t^\tau$, assuming $\mathcal{E}$ is locally $L$-Lipschitz continuous, the deviation induced by the noise term $(1-t)\mathbf{x}_0$ is bounded locally:
$$\|\mathcal{E}(\mathbf{x}_t) - \mathcal{E}(t \mathbf{x}_1)\| \le L(1-t)\|\mathbf{x}_0\|$$

While standard self-attention mechanisms in Diffusion Transformers exhibit unbounded Lipschitz constants, our specific DiT architecture introduces Query-Key Normalization (QK-Norm) as a structural mitigation. By projecting queries and keys onto a unit hypersphere, QK-Norm bounds the attention logits. \textbf{We explicitly acknowledge that this local layer-wise normalization does not provide a strict theoretical global Lipschitz bound for the entire deep DiT backbone.} Instead of claiming a rigorous global mathematical bound, we frame QK-Norm from an empirical optimization perspective: it provides a well-conditioned optimization landscape. By preventing localized exponential amplification of the noise term $(1-t)\mathbf{x}_0$ within the attention maps, this conditioning allows our global empirical regularizer to effectively enforce trajectory-wide noise invariance.

\textbf{3. Empirical Noise Invariance via Consistency Loss}

Given the absence of a strict global Lipschitz bound, neural networks might amplify this noise despite local normalizations. To overcome this empirically, our consistency loss (${\mathcal{L}_{CT\_inv}}$) explicitly penalizes the divergence of the predicted velocity across timesteps:
$$\min_{\theta} \|\mathbf{v}_{inv}(\mathbf{x}_t, t) - \mathbf{v}_{inv}(\mathbf{x}_{t+\Delta t}, t+\Delta t)\|^2$$
Because $\mathbf{x}_t$ and $\mathbf{x}_{t+\Delta t}$ differ precisely by the noise scale $(1-t)$ versus $(1-t-\Delta t)$, driving this difference to zero structurally forces the latent Koopman representation $\mathbf{h}_{inv}$ to become invariant to the noise residual $(1-t)\mathbf{x}_0$. Consequently, the operator $\mathbf{K}$ effectively acts on the expectation of the physical state $\mathbb{E}[\mathbf{x}_t] \propto \mathbf{x}_1$. Rather than implementing a rigorous SKO mathematically, we employ SKO theory as an explanatory framework: it illustrates how mapping dynamics onto the expected state successfully mitigates additive uncertainty. This provides a strong theoretical justification for our architectural design, aligning perfectly with the core principles of SKO without requiring flawless mathematical equivalence.

We explicitly acknowledge that during extreme contact-rich interactions (e.g., rigid collisions or high-frequency disturbances), the local dynamic transition becomes highly non-linear, and this linear Lipschitz bound may momentarily saturate or be violated. However, this limitation forms the exact structural motivation for our dual-branch architecture. The Time-Invariant Koopman operator is only required to bound the macroscopic, low-frequency manifold as a structural prior. When physical contact dynamics exceed this bound, the unconstrained, visually-driven Time-Variant branch ($\mathbf{v}_{var}$) dynamically absorbs the exact residual necessary to satisfy the total target vector field. Thus, global convergence relies not on a single flawless theoretical proof, but on the complementary empirical synergy between a bounded Koopman approximation and unconstrained Flow Matching.

\textbf{4. Bounded Discretization Error in Single-Step Inference}

A critical architectural feature of KoopmanFlow is that during single-step inference ($\Delta t = 1$), the time-variant branch $\mathbf{v}_{var}$ acts as a constant reactive impulse. Standard Flow Matching vector fields often diverge when integrated using a single Euler step over $t \in [0,1]$ because the global vector field $\mathbf{v}_\theta$ varies highly non-linearly. However, in our decoupled architecture, the global trajectory is strongly anchored by the consistency-distilled macroscopic branch $\mathbf{v}_{inv}$. The residual field $\mathbf{v}_{var}$ only models the bounded high-frequency subspace ($\mathbf{x}_{var}$). Because the semantic features driving $\mathbf{v}_{var}$ are conditionally sparse (activated primarily by discrete visual events), the local Lipschitz constant of the variant sub-manifold remains tightly constrained around these event anchors. Mathematically, the single Euler step integration error $E_{Euler}$ is proportional to the temporal gradient of the field:
$$E_{Euler} \propto \sup_{t} \left\| \frac{\partial \mathbf{v}_{var}}{\partial t} \right\| \Delta t^2$$
Since $\mathbf{v}_{var}$ exclusively captures high-frequency spatial residuals rather than the full-state dynamic span, its magnitude $\|\mathbf{v}_{var}\| \ll \|\mathbf{v}_\theta\|$, rendering the single-step discretization error mathematically bounded and empirically stable for reactive RHC deployment.

\vspace{4mm}
\section*{Appendix B: Detailed Formulation of Spatiotemporal Kinematic Constraint ($\mathcal{L}_{reg}$)}
\label{app:reg_loss}

\setcounter{equation}{0}
\renewcommand{\theequation}{B\arabic{equation}}

To ensure the structural stability of the time-invariant Koopman branch discussed in Section 3, we introduce a spatiotemporal kinematic constraint $\mathcal{L}_{reg}$. This regularization is strictly composed of $L_1$ penalty terms to aggressively smooth persistent stochastic diffusion noise while maintaining the capacity to represent intentional macroscopic shifts without over-smoothing. Specifically, $\mathcal{L}_{reg}$ consists of three components corresponding to temporal consistency, acceleration bounding, and spatial smoothing:
\begin{equation}
    \mathcal{L}_{reg} = \mathcal{L}_{temp\_diff} + \mathcal{L}_{change\_rate} + \mathcal{L}_{spatial}
\end{equation}

\textbf{1. Temporal Consistency ($\mathcal{L}_{temp\_diff}$):} We constrain the predictions to be similar across consecutive diffusion timesteps. Let $\mathbf{v}_{inv\_pred}$ be the invariant velocity predicted at the current diffusion step $t$, and $\mathbf{v}_{inv\_next}$ be the EMA teacher's target prediction at $t+\Delta t$:
\begin{equation}
    \mathcal{L}_{temp\_diff} = \lambda_{temporal} \cdot \mathbb{E} \left[ \left\| \mathbf{v}_{inv\_pred} - \mathbf{v}_{inv\_next} \right\|_1 \right]
\end{equation}

\textbf{2. Acceleration Bounding / Change Rate Consistency ($\mathcal{L}_{change\_rate}$):} To mathematically prevent the invariant branch from outputting non-physical, high-frequency acceleration spikes along the temporal action sequence, we penalize the discrepancy in the temporal derivative (change rate along the sequence length $\tau$) between the prediction and the target:
\begin{equation}
    \mathcal{L}_{change\_rate} = \lambda_{temporal} \cdot \mathbb{E} \left[ \left\| \frac{\partial \mathbf{v}_{inv\_pred}}{\partial \tau} - \frac{\partial \mathbf{v}_{inv\_next}}{\partial \tau} \right\|_1 \right]
\end{equation}

\textbf{3. Spatial Smoothing ($\mathcal{L}_{spatial}$):} To bias the invariant Koopman operator toward generalized, stable representations, we constrain the predictions to remain close to the batch mean $\boldsymbol{\mu}_{batch}$, thereby preventing individual samples from exhibiting isolated erratic macroscopic behaviors:
\begin{equation}
    \mathcal{L}_{spatial} = \lambda_{spatial} \cdot \mathbb{E} \left[ \left\| \mathbf{v}_{inv\_pred} - \boldsymbol{\mu}_{batch} \right\|_1 \right]
\end{equation}
By utilizing the absolute error ($L_1$ norm) scaled by hyperparameters $\lambda_{temporal}$ and $\lambda_{spatial}$, the optimization landscape naturally suppresses trivial jittering while selectively tolerating necessary, large-magnitude physical corrections.

\subsection*{B.1 Implementation Details and Hyperparameters}
\textbf{Tikhonov Regularization in Localized DMD:} Regarding the localized Dynamic Mode Decomposition ($DMD_{loc}$) applied in the time-variant branch (Equation 4), the Tikhonov regularization damping factor $\lambda$ serves as a critical hyperparameter. In all our simulated and real-world experiments, we empirically set $\lambda = 10^{-3}$. This specific value provides an optimal trade-off: it is sufficiently large to prevent numerical explosion and manifold degeneration (i.e., preventing the condition number $\kappa(X) \rightarrow \infty$ of the covariance matrix) during abrupt physical singularities such as rigid collisions or sudden grasping impacts. Concurrently, it is small enough to avoid acting as an aggressive low-pass filter, thereby ensuring the precise preservation of high-frequency transient signals necessary for executing visually-driven reactive corrections.

\vspace{4mm}
\section*{Appendix C: Simulation Experiments}
\label{sec:sim_experiment}

To comprehensively evaluate the robustness and generalizability of KoopmanFlow, we conduct extensive experiments in both simulation and real-world settings. This section summarizes the simulation benchmarks, training protocols, configuration details, and qualitative result analysis.

\begin{figure}[H]
    \centering
    \includegraphics[width=\textwidth]{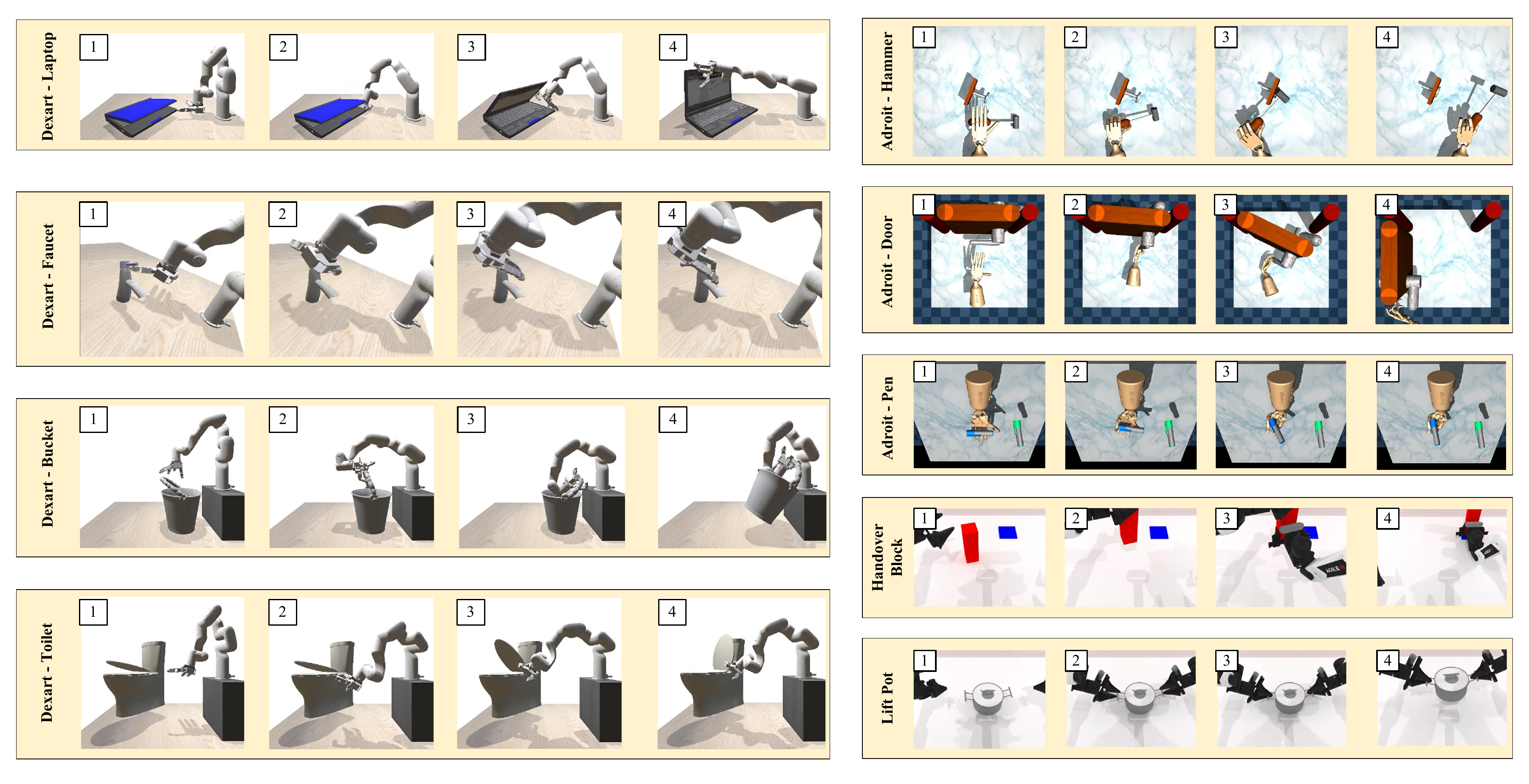}
    \caption{Visualization of representative simulation tasks from Adroit, DexArt, and RoboTwin 2.0.}
    \label{fig:sim_tasks}
\end{figure}

\subsection*{C.1 Benchmarks and Tasks}
\label{subsec:sim_benchmark}

We evaluate KoopmanFlow on four complementary simulation benchmarks that cover diverse manipulation regimes: \textbf{Adroit} for high-DoF dexterous hand control, \textbf{DexArt} for articulated object manipulation, \textbf{MetaWorld} for language-conditioned multi-task learning, and \textbf{RoboTwin 2.0} for contact-rich bimanual coordination. Representative task visualizations are shown in Figure~\ref{fig:sim_tasks}.

\textbf{Adroit.} Adroit focuses on dexterous manipulation with a high-degree-of-freedom robotic hand and requires precise coordination under complex contact dynamics. We report results on \emph{Hammer}, \emph{Door}, and \emph{Pen}.

\textbf{DexArt.} DexArt contains articulated object manipulation tasks involving structured interactions with movable components such as lids, handles, and joints. We evaluate on \emph{Laptop}, \emph{Faucet}, \emph{Bucket}, and \emph{Toilet}.

\textbf{MetaWorld.} For MetaWorld, we follow the standard 48-task benchmark and focus on the language-conditioned multi-task setting, which provides a more challenging evaluation protocol than single-task training.

\textbf{RoboTwin 2.0.} RoboTwin 2.0 emphasizes robust dual-arm coordination and contact-rich interaction. We evaluate on five representative tasks: \emph{Adjust Bottle}, \emph{Press Stapler}, \emph{Lift Pot}, \emph{Stack Bowls 3}, and \emph{Handover Block}.

\subsection*{C.2 Training and Evaluation Protocol}
\label{subsec:sim_training}

For fair comparison, we follow the same simulation training protocol as ManiFlow. Following prior benchmark-specific settings, we collect different amounts of expert demonstrations according to task complexity. The number of demonstrations used for each benchmark is summarized in Table~\ref{tab:sim_demo_counts}.

\begin{table}[H]
    \centering
    \caption{Number of expert demonstrations used in each simulation benchmark.}
    \label{tab:sim_demo_counts}
    \begin{tabular}{lc}
        \toprule
        \textbf{Benchmark} & \textbf{Demonstrations (Per Task)} \\
        \midrule
        Adroit & 10 \\
        DexArt & 100  \\
        MetaWorld & 10  \\
        RoboTwin 2.0 & 50 \\
        \bottomrule
    \end{tabular}
\end{table}

\textbf{Adroit and DexArt.} All models are trained for 3000 epochs. Evaluation is performed every 50 epochs over 20 episodes, and the reported results are computed as the average of the top five success rates across evaluated checkpoints in order to reduce checkpoint-level variance.

\textbf{MetaWorld.} Given that many robust baselines have already reached performance ceilings across most single-task environments in MetaWorld, our evaluation shifts away from isolated tasks. Instead, we target the significantly more demanding language-guided multi-task setting. The training and evaluation protocol follows the same setup as Adroit and DexArt.

\textbf{RoboTwin 2.0.} All models are trained for 2000 epochs. We evaluate the final checkpoint over 100 episodes following the standard benchmark protocol. Across all benchmarks, model effectiveness and consistency are quantified by documenting the mean success rates and standard deviations, which are derived from three separate training runs.

\subsection*{C.3 Policy Configuration}
\label{subsec:sim_config}

We use action horizons of 4 steps and an observation history of 2 timesteps for the simulation tasks in \textbf{Adroit}, \textbf{DexArt}, and \textbf{MetaWorld}. During rollout, the policy predicts a 4-step action sequence and executes the first 3 steps before replanning.

For dexterous bimanual tasks in \textbf{RoboTwin 2.0}, we use a longer action horizon of 16 steps together with an observation history of 8 timesteps. During rollout, the policy predicts a 16-step action sequence and executes the first 12 steps before replanning. This setting is adopted to better accommodate the longer-horizon coordination patterns and denser contact transitions required in bimanual manipulation.

\subsection*{C.4 Simulation Result Analysis}
\label{subsec:sim_analysis}

\begin{figure}[H]
    \centering
    \includegraphics[width=\textwidth]{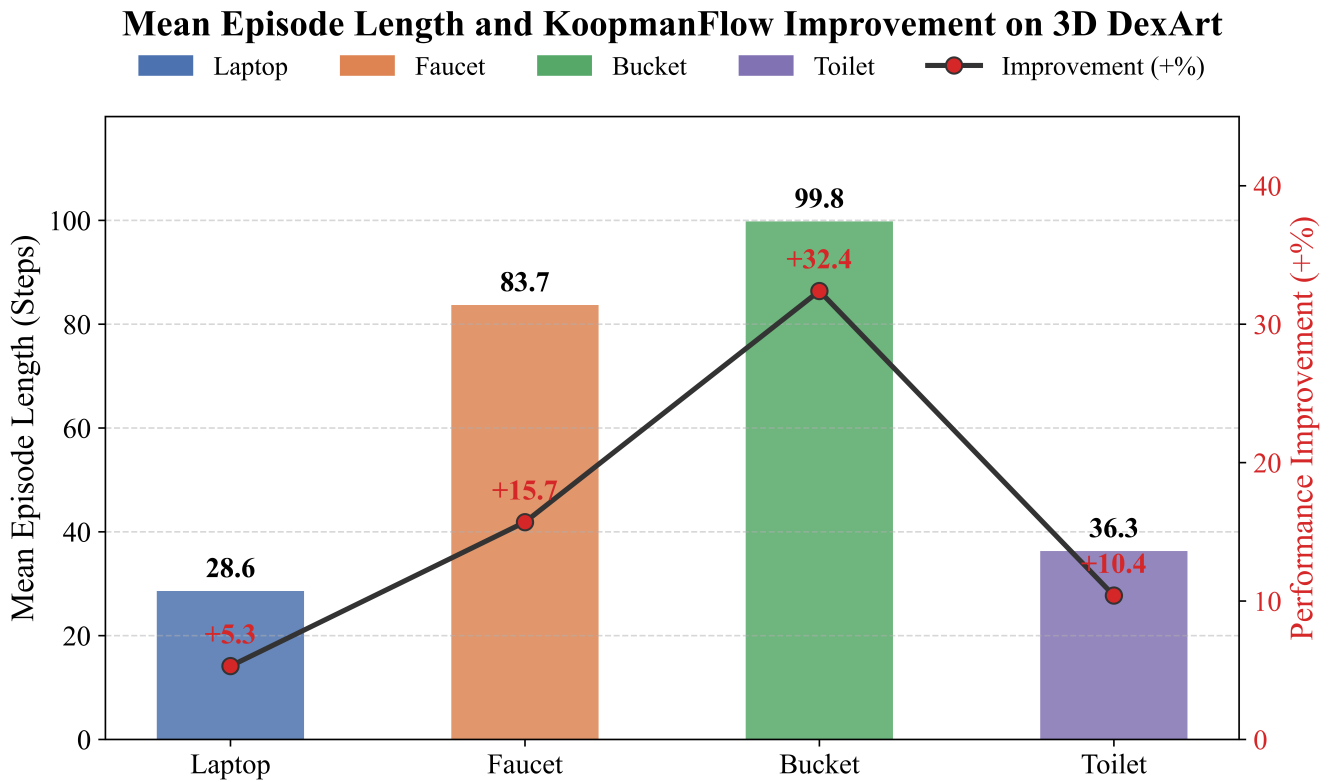}
    \caption{Mean episode length and relative performance improvement of KoopmanFlow on 3D DexArt. Tasks with longer effective horizons generally show larger gains, supporting the claim that KoopmanFlow is particularly beneficial for long-horizon and contact-rich manipulation.}
    \label{fig:dexart_length_improve}
\end{figure}

\textbf{Successful cases.} We analyze simulation results from two complementary perspectives: \emph{macro-trajectory stability} and \emph{transient correction quality}. Successful cases typically show that KoopmanFlow preserves a stable global motion trend while remaining responsive to task-critical local events. Under visual conditioning, the policy captures informative cues and performs localized corrections at key stages of manipulation, thereby improving overall execution success. As Figure~\ref{fig:dexart_length_improve} shows, this advantage becomes particularly evident in long-horizon and contact-rich tasks, where local corrective behavior helps mitigate \emph{compound error accumulation}.

\textbf{Failure cases.} For 2D simulation, failures are more likely to arise from the joint effect of \emph{visual degradation} and \emph{contact transient difficulty}. Since 2D observations are more susceptible to partial occlusion, ambiguous viewpoints, or degraded visual cues, the policy may fail to reliably capture the critical events required for precise local correction. These perceptual limitations, together with the inherent difficulty of long-horizon and contact-rich manipulation, can lead to unsuccessful executions. For 3D simulation, failure cases are generally better explained by \emph{dynamics-side difficulty}. In particular, residual failures often occur in long-horizon or contact-rich tasks, where small deviations accumulate over time and eventually lead to unsuccessful completion. In some cases, abrupt contact transitions or highly non-stationary interaction patterns still require stronger transient corrections than the current one-step policy can provide. Nevertheless, across these challenging settings, KoopmanFlow consistently achieves better success rates than all compared baselines.

\section*{Appendix D: Real-World Experiments}
\label{sec:real_experiment}

\subsection*{D.1 Experimental Setup and Tasks}
\label{sec:real_setup}
\begin{figure}[H]
    \centering
    \includegraphics[width=0.9\textwidth]{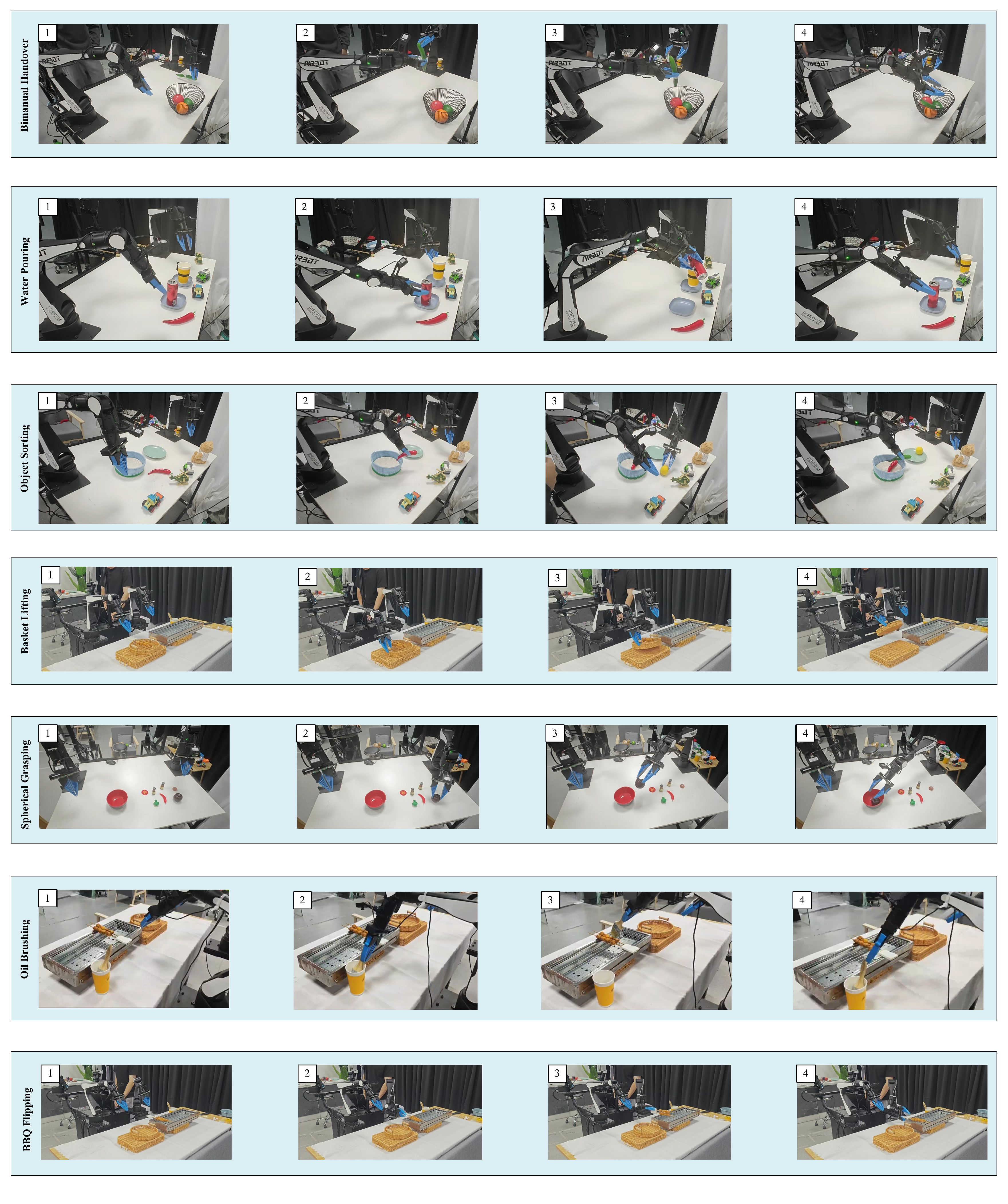}
    \caption{Visualization of Real-World Tasks. The tasks span from complex bimanual coordination to high-precision single-arm interactions.}
    \label{fig:real_tasks}
\end{figure}

Beyond simulation, we deploy KoopmanFlow on real-world setups to evaluate its generalizability across both single-arm and bimanual configurations. Figure~\ref{fig:real_tasks} illustrates the execution trajectories of our real-world tasks.

The real-world evaluation encompasses the following tasks, divided by their hardware configurations:

\textbf{Bimanual Tasks:}
\begin{itemize}
     \item \textbf{Bimanual Handover}: One arm grasps an object from the workspace and safely hands it over to the other arm mid-air, which then deposits the item into a basket in front of the robot.
    \item \textbf{Water Pouring}: One arm lifts and stabilizes a target container while the other arm precisely pours water from a bottle into it; subsequently, the arms independently place the bottle down and present the filled container forward.
    \item \textbf{Object Sorting}: The dual arms must identify distinct objects and simultaneously sort them into their respective target containers.
\end{itemize}

\textbf{Single-Arm Tasks:}
\begin{itemize}
    \item \textbf{Basket Lifting}: The robot grasps the handle of a basket and lifts it stably without dropping the contained items.
    \item \textbf{Spherical Grasping}: The gripper must accurately locate and firmly grasp spherical objects scattered on the workspace, demonstrating robustness against tabletop visual distractors.
    \item \textbf{Oil Brushing}: The arm retrieves a brush from a container, applies oil smoothly across a designated surface, and returns the brush to its original container.
    \item \textbf{BBQ Flipping}: The robot end-effector must precisely locate and flip a BBQ skewer on a grill.
\end{itemize}

\subsection*{D.2 Real-World Results Analysis}
\label{sec:real_results}

\textbf{Successful cases.} We analyze the real-world results from the perspectives of \emph{global trajectory stability} and \emph{local reactive correction}. Successful executions typically indicate that the time-invariant branch helps preserve a smooth and stable overall manipulation trajectory, while the time-variant branch provides timely local corrections during grasping, contact, and external perturbations. This division of roles improves execution robustness and contributes to the consistently stronger task success rates of KoopmanFlow in real-world settings.

\begin{figure}[H]
    \centering
    \includegraphics[width=\textwidth]{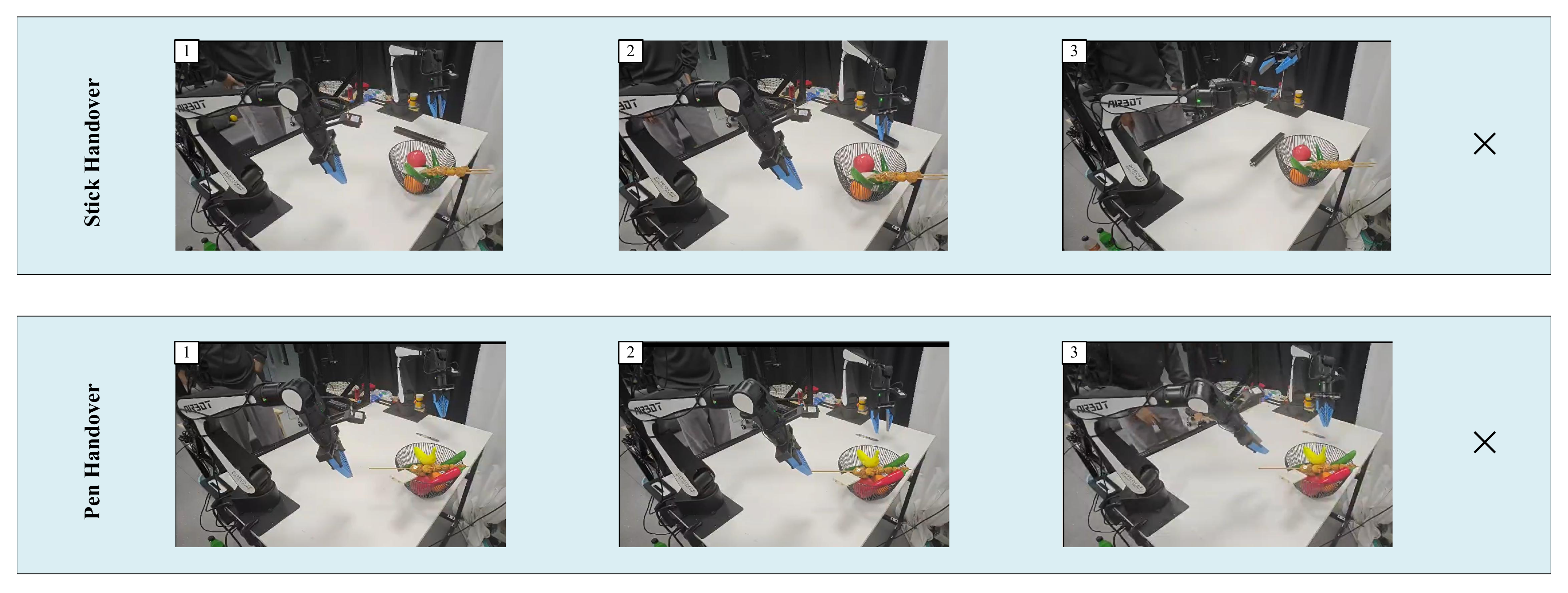}
    \caption{Representative failure cases in real-world unseen-object handover tasks. The top row shows Stick Handover, where the manipulated object is substantially larger and heavier than those seen during training, making the contact-transfer dynamics harder to handle and leading to unsuccessful execution. The bottom row shows Pen Handover, where the small object size together with finger-object occlusion increases the difficulty of precise grasp localization, causing failure at an earlier grasping stage.}
    \label{fig:failure-realworld}
\end{figure}

\textbf{Failure cases.} In real-world scenarios, failure cases are more likely to arise from the combined effect of \emph{visual uncertainty} and \emph{contact dynamics complexity}. Since the policy relies solely on monocular 2D observations, perception can be more easily affected by viewpoint ambiguity, partial occlusion, appearance variation, or degraded visual cues. At the same time, real-world contact dynamics are often more complex and less predictable than those in simulation, making accurate transient correction more difficult. As a result, small errors introduced during perception or early-stage interaction may accumulate over time, eventually causing instability in the later stage of the task and leading to failure.

Figure~\ref{fig:failure-realworld} shows two representative failure cases from unseen-object handover tasks. In the top example (Stick Handover), the stick is substantially larger and heavier than the objects seen during training. This changes the contact and transfer dynamics during handover, making the transient interaction harder to model accurately and ultimately leading to unsuccessful execution. In the bottom example (Pen Handover), the pen is much smaller, which makes it more vulnerable to localization errors and finger-object occlusion during grasp closure. As a result, the failure occurs earlier, at the grasping stage, before a stable handover can be completed.

\subsection*{D.3 Real-World Behavior under Instruction--Object Mismatch}
\label{sec:real_identify}

\begin{figure}[htbp]
    \centering
    \includegraphics[width=\textwidth]{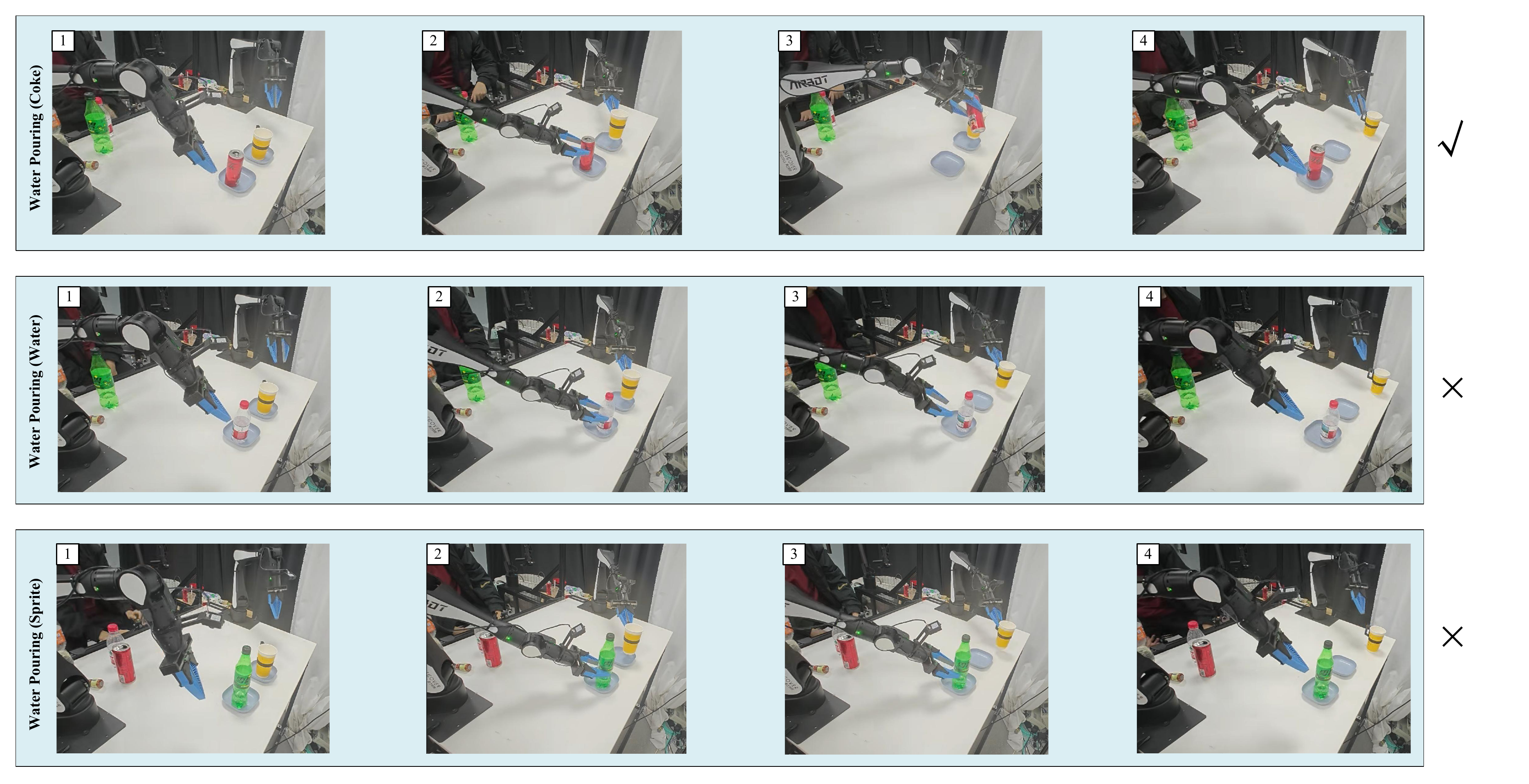}
    \caption{Water Pouring under instruction-object mismatch. Each row shows a four-step sequence. When the bottle identity matches the instruction (Coke), the policy completes the pouring task successfully. When the instruction is kept as Coke but the manipulated bottle is mismatched (Water or Sprite), the evolving monocular observation becomes inconsistent with the language-conditioned target identity during manipulation, and the policy suppresses the lift-and-pour transient and transitions to a conservative termination routine.}
    \label{fig:pouring_mismatch}
\end{figure}

As Figure~\ref{fig:pouring_mismatch} shows, we further examine KoopmanFlow under instruction-object mismatch in real-world deployment. The policy exhibits an unanticipated yet structurally coherent fallback behavior when the language instruction conflicts with the manipulated object. When we intentionally introduce a mismatch—for example, by issuing the instruction Coke while presenting a Sprite bottle—the policy can still initiate the reaching and grasping motion. However, as the object pose evolves during manipulation, the monocular visual observation provides identity-specific cues that become inconsistent with the language condition. Instead of continuing the planned lift-and-pour sequence, the policy suppresses the high-frequency pouring transient and transitions to a smooth and conservative fallback routine. Concretely, the right arm releases and retracts without executing the pouring motion, while the left arm continues its stabilizing adjustments for the receiving container.

We view this behavior as being consistent with the spectral decoupling mechanism enforced by the terminal Hybrid Koopman FFN. In standard monolithic policies, a visual-semantic mismatch may perturb the entire continuous vector field and lead to unstable global motion. In KoopmanFlow, by contrast, the macroscopic branch structurally anchors slow-varying trajectories ($v_{inv}$), while the transient branch models high-frequency residuals driven by precise visual cues ($v_{var}$). When the observed object becomes inconsistent with the language condition, the stimulus required to trigger the intended pouring transient is no longer supported by the visual stream. As a result, the right-arm response is dominated by the time-invariant branch, yielding a smoother and physically plausible fallback behavior under distribution shift.
\vspace{4mm}
\begin{table}[H]
\centering
\newcommand{\tabincell}[2]{\begin{tabular}{@{}#1@{}}#2\end{tabular}}
\caption{\textbf{Language-conditioned multi-task results on 48 Meta-World simulation tasks.} }
\vspace{2mm}

\resizebox{\textwidth}{!}{%
\begin{minipage}{1.2\textwidth} 
\centering

\begin{tabular*}{\linewidth}{l@{\extracolsep{\fill}}ccccccc}
\toprule
& \multicolumn{7}{c}{\textbf{Meta-World (Easy)}} \\
Alg $\backslash$ Task &
\tabincell{c}{Button \\ Press} &
\tabincell{c}{Button Press \\ Topdown} &
\tabincell{c}{Button Press \\ Topdown Wall} &
\tabincell{c}{Button Press \\ Wall} &
\tabincell{c}{Coffee \\ Button} &
\tabincell{c}{Dial \\ Turn} &
\tabincell{c}{Door \\ Close} \\
\midrule
3D Diffusion Policy & $62\pm15$ & $100\pm0$ & $100\pm0$ & $72\pm25$ & $73\pm34$ & $53\pm15$ & $100\pm0$ \\
3D Flow Matching & $0\pm0$   & $100\pm0$ & $100\pm0$ & $67\pm31$ & $97\pm5$  & $70\pm7$  & $100\pm0$ \\
3D ManiFlow         & $100\pm0$ & $100\pm0$ & $100\pm0$ & $100\pm0$ & $100\pm0$ & $67\pm13$ & $100\pm0$ \\
\textbf{KoopmanFlow}   & $\mathbf{100\pm0}$ & $\mathbf{100\pm0}$ & $\mathbf{100\pm0}$ & $\mathbf{100\pm0}$ & $\mathbf{100\pm0}$ & $\mathbf{96\pm2}$ & $\mathbf{100\pm0}$ \\
\bottomrule
\end{tabular*}

\vspace{3mm}

\begin{tabular*}{\linewidth}{l@{\extracolsep{\fill}}ccccccc}
\toprule
& \multicolumn{7}{c}{\textbf{Meta-World (Easy)}} \\
Alg $\backslash$ Task & Door Lock & Door Open & Door Unlock & Drawer Close & Drawer Open & Faucet Close & Faucet Open \\
\midrule
3D Diffusion Policy & $0\pm0$ & $100\pm0$ & $98\pm2$ & $88\pm13$ & $98\pm2$ & $92\pm8$ & $83\pm12$ \\
3D Flow Matching & $0\pm0$ & $100\pm0$ & $100\pm0$ & $5\pm7$   & $100\pm0$ & $92\pm6$ & $100\pm0$ \\
3D ManiFlow         & $78\pm14$ & $100\pm0$ & $100\pm0$ & $100\pm0$ & $100\pm0$ & $100\pm0$ & $100\pm0$ \\
\textbf{KoopmanFlow}   & $\mathbf{86\pm4}$ & $\mathbf{100\pm0}$ & $\mathbf{100\pm0}$ & $\mathbf{100\pm0}$ & $\mathbf{100\pm0}$ & $\mathbf{100\pm0}$ & $\mathbf{100\pm0}$ \\
\bottomrule
\end{tabular*}

\vspace{3mm}

\begin{tabular*}{\linewidth}{l@{\extracolsep{\fill}}cccccc}
\toprule
& \multicolumn{6}{c}{\textbf{Meta-World (Easy)}} \\
Alg $\backslash$ Task & Handle Press & Handle Pull & Handle Pull Side & Lever Pull & Plate Slide & Plate Slide Back \\
\midrule
3D Diffusion Policy & $100\pm0$ & $22\pm17$ & $43\pm6$ & $60\pm12$ & $20\pm18$ & $92\pm12$ \\
3D Flow Matching & $100\pm0$ & $15\pm18$ & $20\pm7$ & $45\pm11$ & $0\pm0$   & $88\pm10$ \\
3D ManiFlow         & $100\pm0$ & $42\pm10$ & $65\pm7$ & $63\pm19$ & $100\pm0$ & $93\pm9$ \\
\textbf{KoopmanFlow}   & $\mathbf{100\pm0}$ & $\mathbf{77\pm6}$ & $\mathbf{72\pm6}$ & $\mathbf{85\pm4}$ & $\mathbf{100\pm0}$ & $\mathbf{98\pm2}$ \\
\bottomrule
\end{tabular*}

\vspace{3mm}

\begin{tabular*}{\linewidth}{l@{\extracolsep{\fill}}cccc}
\toprule
& \multicolumn{4}{c}{\textbf{Meta-World (Easy)}} \\
Alg $\backslash$ Task & Plate Slide Back Side & Plate Slide Side & Reach & Reach Wall \\
\midrule
3D Diffusion Policy & $100\pm0$ & $92\pm6$ & $48\pm2$ & $25\pm4$ \\
3D Flow Matching & $100\pm0$ & $82\pm14$ & $57\pm10$ & $35\pm11$ \\
3D ManiFlow         & $100\pm0$ & $100\pm0$ & $58\pm19$ & $67\pm9$ \\
\textbf{KoopmanFlow}   & $\mathbf{100\pm0}$ & $\mathbf{100\pm0}$ & $\mathbf{60\pm6}$ & $\mathbf{76\pm5}$ \\
\bottomrule
\end{tabular*}

\vspace{3mm}

\begin{tabular*}{\linewidth}{l@{\extracolsep{\fill}}ccccccc}
\toprule
& \multicolumn{7}{c}{\textbf{Meta-World (Medium)}} \\
Alg $\backslash$ Task & Window Close & Window Open & Basketball & Bin Picking & Box Close & Coffee Pull & Coffee Push \\
\midrule
3D Diffusion Policy & $100\pm0$ & $83\pm17$ & $100\pm0$ & $0\pm0$ & $18\pm8$ & $52\pm23$ & $55\pm0$ \\
3D Flow Matching & $100\pm0$ & $92\pm6$  & $90\pm4$  & $18\pm6$ & $18\pm8$ & $67\pm2$  & $58\pm15$ \\
3D ManiFlow         & $100\pm0$ & $97\pm2$  & $100\pm0$ & $33\pm2$ & $45\pm12$ & $97\pm2$  & $82\pm6$ \\
\textbf{KoopmanFlow}   & $\mathbf{100\pm0}$ & $\mathbf{100\pm0}$ & $\mathbf{100\pm0}$ & $\mathbf{34\pm6}$ & $\mathbf{65\pm6}$ & $\mathbf{100\pm0}$ & $\mathbf{83\pm4}$ \\
\bottomrule
\end{tabular*}

\vspace{3mm}

\begin{tabular*}{\linewidth}{l@{\extracolsep{\fill}}ccccccc}
\toprule
& \multicolumn{7}{c}{\textbf{Meta-World (Medium)}} \\
Alg $\backslash$ Task & Peg Unplug Side & Hammer & Peg Insert Side & Push Wall & Soccer & Sweep & Sweep Into \\
\midrule
3D Diffusion Policy & $60\pm7$  & $77\pm6$  & $58\pm6$  & $60\pm8$  & $8\pm5$  & $70\pm4$ & $3\pm2$ \\
3D Flow Matching & $68\pm5$  & $75\pm22$ & $68\pm5$  & $15\pm15$ & $10\pm4$ & $63\pm21$ & $0\pm0$ \\
3D ManiFlow         & $80\pm2$  & $42\pm24$ & $88\pm8$  & $93\pm2$  & $7\pm6$  & $92\pm2$ & $7\pm2$ \\
\textbf{KoopmanFlow}   & $\mathbf{85\pm5}$ & $\mathbf{78\pm5}$ & $\mathbf{100\pm0}$ & $\mathbf{97\pm2}$ & $\mathbf{17\pm4}$ & $\mathbf{96\pm2}$ & $\mathbf{52\pm7}$ \\
\bottomrule
\end{tabular*}

\vspace{3mm}

\begin{tabular*}{\linewidth}{l@{\extracolsep{\fill}}ccccc}
\toprule
& \multicolumn{5}{c}{\textbf{Meta-World (Hard)}} \\
Alg $\backslash$ Task & Assembly & Hand Insert & Pick Out of Hole & Pick Place & Push \\
\midrule
3D Diffusion Policy & $77\pm16$ & $7\pm9$ & $20\pm11$ & $42\pm5$ & $55\pm14$ \\
3D Flow Matching & $88\pm10$ & $0\pm0$ & $ \mathbf{38\pm2}$  & $53\pm17$ & $62\pm2$ \\
3D ManiFlow         & $100\pm0$ & $12\pm9$ & $13\pm5$ & $\mathbf{68\pm5}$ & $88\pm9$ \\
\textbf{KoopmanFlow}   & $\mathbf{100\pm0}$ & $\mathbf{14\pm4}$ & $35\pm6$ & $65\pm5$ & $\mathbf{90\pm3}$ \\
\bottomrule
\end{tabular*}

\vspace{3mm}

\begin{tabular*}{\linewidth}{l@{\extracolsep{\fill}}ccccc|c}
\toprule
& \multicolumn{5}{c|}{\textbf{Meta-World (Very Hard)}} & \textbf{Average} \\
Alg $\backslash$ Task & Shelf Place & Disassemble & Stick Pull & Stick Push & Pick Place Wall & \textbf{Avg.} \\
\midrule
3D Diffusion Policy & $25\pm8$  & $55\pm19$ & $28\pm14$ & $85\pm3$  & $55\pm27$ & $59.4 \pm 3.5$ \\
3D Flow Matching & $18\pm10$ & $67\pm5$  & $43\pm25$ & $88\pm5$  & $40\pm11$ & $57.9 \pm 0.5$ \\
3D ManiFlow         & $28\pm5$  & $63\pm8$  & $83\pm5$  & $95\pm5$  & $98\pm2$  & $78.1 \pm 2.0$ \\
\textbf{KoopmanFlow}   & $\mathbf{34\pm6}$ & $\mathbf{81\pm4}$ & $\mathbf{90\pm0}$ & $\mathbf{100\pm0}$ & $\mathbf{100\pm0}$ & $\mathbf{84.7\pm1.5}$ \\
\bottomrule
\end{tabular*}

\end{minipage}
}
\end{table}

\vspace{4mm}
\section*{Appendix E: Pseudocode for KoopmanFlow}
\label{sec:pseudocode}

To provide a concrete algorithmic overview of our proposed architecture and facilitate reproducibility, we present the condensed pseudocode for the training and inference pipelines of KoopmanFlow. The algorithms abstract standard diffusion routines to focus strictly on our core structural contributions.

\subsection*{E.1 Fused Co-Training Pipeline}
Algorithm~\ref{alg:training} outlines the Fused Co-Training process. We mitigate gradient conflicts between Optimal Transport Flow Matching (OT-FM) and consistency distillation by introducing an asymmetric spatial batch partitioning strategy. Within each training iteration, the batch is divided into two distinct subsets based on the ratio $r_{ct}$. The majority subset is dedicated to anchoring the continuous global trajectory via the CFM objective, while the smaller subset undergoes single-step consistency distillation.

During the forward pass, the latent trajectory representations are spectrally decoupled via a 1D Real Fast Fourier Transform (RFFT) guided by a cumulative energy threshold $\alpha$. The macroscopic Stochastic Koopman Operator (SKO) processes the low-frequency components to capture stable physical inertia. The high-frequency components are handled by a localized Dynamic Mode Decomposition ($DMD_{loc}$) over a sliding window $\tau_w$ to isolate visually-driven reactive transients. The network is ultimately optimized using the unified objective across the partitioned batch.

\begin{algorithm}[H]
\caption{Fused Co-Training of KoopmanFlow}
\label{alg:training}
\begin{algorithmic}[1]
\Require Training dataset $\mathcal{D}$, partition ratio $r_{ct}$, energy threshold $\alpha$
\For{each training iteration}
    \State Sample batch $(X_1, c_{obs}) \sim \mathcal{D}$
    \State Sample $t \sim \mathcal{U}(0,1)$, $X_0 \sim \mathcal{N}(0, I)$, and compute $X_t = tX_1 + (1-t)X_0$
    \State Split batch indices into $I_{CT}$ (size $B \cdot r_{ct}$) and $I_{FM}$ (size $B \cdot (1-r_{ct})$)
    \State \textit{\# Spectrally Decoupled Forward Pass}
    \State $h_{in} = \text{DiTBlocks}_\theta(X_t, c_{obs}, t)$
    \State $X_{inv}, X_{var} = \text{RFFT-Filter}(h_{in}, \alpha)$
    \State $v_{inv} = \text{Proj}_{inv}(\text{SKO}(X_{inv}, \tilde{K}_{inv}))$ \hfill $\triangleright$ Anchor macroscopic inertia
    \State $v_{var} = \text{Proj}_{var}(\text{DMD}_{loc}(X_{var}, \tau_w))$ \hfill $\triangleright$ Isolate reactive transients
    \State $v_\theta = v_{inv} + v_{var}$
    \State \textit{\# Asymmetric Objective Optimization}
    \State $\mathcal{L}_{FM} = \mathbb{E}_{I_{FM}} \left[ ||v_\theta - (X_1 - X_0)||_2^2 \right]$
    \State $\mathcal{L}_{CT} = \text{ConsistencyDistillation}(v_\theta[I_{CT}], \theta_{EMA})$
    \State $\mathcal{L}_{dec} = \text{ExplicitDecoupling}(v_{inv}[I_{CT}], v_{var}[I_{CT}], \theta_{EMA})$
    \State Update $\theta$ using $\nabla_\theta (\mathcal{L}_{FM} + \mathcal{L}_{CT} + \mathcal{L}_{dec} + \mathcal{L}_{reg})$ and step $\theta_{EMA}$
\EndFor
\end{algorithmic}
\end{algorithm}

\subsection*{E.2 Real-Time 1-Step Inference for Deployment}
Algorithm~\ref{alg:inference} shows the deployment execution of KoopmanFlow under real-time Receding Horizon Control (RHC) constraints. By collapsing the macroscopic generation trajectory during training, the model achieves high-fidelity action generation in a single Neural Function Evaluation ($NFE = 1$). The network predicts the decoupled velocity fields directly at $t=0$, which are used to project the noise manifold to the clean physical sequence via a single Euler integration step. The policy then extracts the immediate action step for execution and replans.

\begin{algorithm}[H]
\caption{1-Step Action Generation (RHC Deployment)}
\label{alg:inference}
\begin{algorithmic}[1]
\Require Policy $\theta$, current state $o_\tau$, semantic instruction $l$
\State $c_{obs} = \text{Encode}(o_\tau, l)$
\State Sample $X_0 \sim \mathcal{N}(0, I)$, set $t=0, \Delta t=1$ \hfill $\triangleright$ Initialize for NFE = 1
\State $h_{in} = \text{DiTBlocks}_\theta(X_0, c_{obs}, 0)$
\State $X_{inv}, X_{var} = \text{RFFT-Filter}(h_{in}, \alpha)$
\State $v_\theta = \text{Proj}_{inv}(\text{SKO}(X_{inv})) + \text{Proj}_{var}(\text{DMD}_{loc}(X_{var}))$
\State $X_1 = X_0 + v_\theta \cdot \Delta t$ \hfill $\triangleright$ Single Euler step integration
\State \textbf{return} $X_1[0]$ \hfill $\triangleright$ Execute the first action step $a_\tau$
\end{algorithmic}
\end{algorithm}
\end{document}